\documentclass[comsoc,journal]{IEEEtran} 
\usepackage{tablefootnote}
\usepackage{float}
\usepackage{cite}
\usepackage{setspace}
\usepackage{graphicx}
\usepackage{newtxmath}
\usepackage{hyperref}
\usepackage{amsfonts}
\usepackage{amsmath}
\usepackage{algorithm,algcompatible}
\usepackage{fancyhdr}
\usepackage{xcolor}
\usepackage{multirow}
\usepackage{flafter}
\usepackage{url}
\usepackage{enumerate}
\usepackage{subfigure} 
\usepackage{epstopdf}
\usepackage{booktabs}
\usepackage{multirow}
\providecommand{\keywords}[1]{\textbf{\textit{Index terms---}} #1}

\begin{document}
\title{Federated Prompt-based Decision Transformer for Customized VR Services in Mobile Edge Computing System}

\author{Tailin~Zhou, \IEEEmembership{Graduate Student Member,~IEEE,}
        Jiadong~Yu,\IEEEmembership{ Member,~IEEE,} 
        Jun~Zhang,~\IEEEmembership{Fellow,~IEEE,}
        and~Danny~H.K.~Tsang,~\IEEEmembership{Life~Fellow,~IEEE}
\thanks{

T. Zhou is with IPO,  Academy of Interdisciplinary Studies, The Hong Kong University of Science and Technology, Clear Water Bay, Hong Kong SAR, China (Email: tzhouaq@connect.ust.hk).

J. Yu is with the Internet of Things Thrust, The Hong Kong University of Science and Technology (Guangzhou), Guangzhou, Guangdong, China (Email: jiadongyu@hkust-gz.edu.cn). 

J. Zhang is with the Department of Electronic and Computer Engineering, The Hong Kong University of Science and Technology, Clear Water Bay, Hong Kong SAR, China  (E-mail: eejzhang@ust.hk).

D. H.K. Tsang is with the Internet of Things Thrust, The Hong Kong University of Science and Technology (Guangzhou), Guangzhou, Guangdong, China, and also with the Department of Electronic and Computer Engineering, The Hong Kong University of Science and Technology, Clear Water Bay, Hong Kong SAR, China (Email: eetsang@ust.hk). 
}
}
\maketitle
\begin{abstract}
This paper investigates resource allocation to provide heterogeneous users with customized virtual reality (VR) services in a mobile edge computing (MEC) system. 
We first introduce a quality of experience (QoE) metric to measure user experience, which considers the MEC system's latency, user attention levels, and preferred resolutions.
Then, a QoE maximization problem is formulated for resource allocation to ensure the highest possible user experience,  
which is cast as a reinforcement learning problem, aiming to learn a generalized policy applicable across diverse user environments for all MEC servers.
To learn the generalized policy,  we propose a framework that employs federated learning (FL) and prompt-based sequence modeling to pre-train a common decision model across MEC servers, which is named FedPromptDT.
Using FL solves the problem of insufficient local MEC data while protecting user privacy during offline training.
The design of prompts integrating user-environment cues and user-preferred allocation improves the model's adaptability to various user environments during online execution.
 Through extensive experimental evaluations, we demonstrate that FedPromptDT outperforms baseline methods and exhibits remarkable adaptability, maintaining superior performance across various user environments.
\end{abstract}


\section{Introduction}
\label{sec:intro}

There has been great interest in the metaverse, which provides participants with a deeply engaging and interactive virtual environment \cite{WangSZXLLS23Metaverse}. 
One of its applications is the virtual reality (VR) service, which is highly sensitive to timing.  
Increased latency between input and display in VR services can significantly harm the quality of experience (QoE) for users, resulting in a range of discomforts, from mild unease to severe motion sickness, with symptoms such as disorientation, nausea, and vomiting \cite{ChangKY20VRSickness}. 
To reduce this latency, one potential solution is using the mobile edge computing (MEC) system \cite{MaoYZHL17} for infrastructure. 
The MEC system can offload resource-intensive tasks to edge servers \cite{XuNLKXNYSM23,yu2022bi}, e.g., reducing the time to render graphics and transmit rendered data.
An illustrative example is the field of view (FoV) processing for 360° VR video \cite{yu2023attention}. 
The MEC servers process incoming user interactions, render the  FoV, and then stream the rendered content back to the user's VR devices for real-time viewing.

 Improving users' immersive experience for VR streaming can also provide a better QoE for users.
 Attention-aware rendering  \cite{8717893,8626197}  is a visual-attention-based method for prioritizing high-quality rendering in the regions where the VR user's eyes are focused, making it particularly effective for immersive experiences. 
This is because human vision attention operates hierarchically and selectively, prioritizing certain regions within the FoV with varying degrees of clarity. 
The field of gaze attention, where the viewer is looking directly, is the region with the highest clarity.
With attention-aware rendering, the MEC servers can allocate more resources to the user's focused regions based on the human eye's attention hierarchy to enhance the user's QoE while reducing the rendering requirements for VR streaming.

Furthermore, emphasizing user preferences can improve users' QoE since individual users may have varying priorities regarding their viewing experience.  
This user-centric flexibility ensures users' immersive experience.
For instance, some users may prioritize achieving a higher resolution, ensuring a seamless and realistic virtual environment, while others may lean towards a smoother streaming experience with appropriate resolution. 
Thus, MEC servers can offer a spectrum of resolution levels for users to tailor their settings according to their diverse preferences and device capabilities. 

Motivated by the above user-centric needs (i.e., visual attention and user preferences),  we introduce a new QoE metric for individual users to quantify their experience with the service provided by the MEC system.
The QoE metric incorporates the MEC system's latency, user attention levels, and preferred resolutions to measure users' immersive experiences. 
We optimize resource allocation for MEC servers to maximize the overall QoE and ensure users' immersive experience.
Due to varying user preferences and dynamic communication conditions, defining the environment of MEC systems in advance can be challenging.
This leads to the allocation problem that cannot be solved using explicit optimization policies.
Hence, we transform the problem into a reinforcement learning (RL) problem. 
The goal is to learn a generalized policy that can be applied to various user environments across all MEC servers. 

However, existing value policy-based RL methods typically learn sub-optimal policies that lack generalization capabilities for diverse environments as per \cite{li2023RLTransformers}.
When applied to our RL problem, these methods necessitate re-training or online fine-tuning to tackle diverse user environments not encountered during their training.
Recently, with the emergence of generative pre-trained Transformer (GPT) \cite{radford2018improving_GPT1,Brown2020GPT3}, transformer models showcase a unique capacity for generalization, enabling them to make accurate predictions with a few shots of demonstrations in natural language processing tasks.
Inspired by the transformer structure, the Decision Transformer (DT) \cite{Chen2021decision} is an innovative approach that re-frames RL as a sequence modeling problem conditioned by a desired return.
Generative trajectory modeling helps the DT generate actions under the desired outcome by predicting the patterns of states, actions, and rewards.

 We propose a federated prompt-based decision transformer (FedPromptDT) framework, which exhibits a strong generalization capability in diverse user environments across all MEC servers.
This framework utilizes DT architectures with prompt-based capabilities for learning a generalized policy in our RL problem.
However, training a prompt-based decision transformer (PromptDT) model requires various trajectory data to handle customized user requirements across MEC servers. 
Limited data availability at each MEC server or strict privacy requirements for centralized cloud training may prevent the direct application of this approach.
Our FedPromptDT framework thus employs federated learning (FL) \cite{mcmahan2017communication} to pre-train a PromptDT model across MEC servers based on their local data, where the PromptDT model trained by FL is denoted as FedPromptDT in this work.
Moreover,  our prompt-based method incorporates the information on user environments and user-preferred allocation to prompt the FedPromptDT model to generate optimal resource allocation for MEC servers. 
This overcomes the generalization problems of existing RL methods in our scenario and eliminates the need for retraining or online-tuning the model during online execution.

\begin{table}[t]
    \centering
\caption{List of the notations and their definition. The subscript $e$ and $k$ refer to the $e$-th MEC server and the $k$-th user, respectively.}
\label{tab:notation_list}
\scalebox{0.8}
{
    \begin{tabular}{ll}
    \toprule 
        Notations & Definition \\ \midrule   
        $e, E$ & MEC server index, total number of MEC servers \\
        $k, K_e$  & User index, total number of users\\ 
        $t, R$ & Time index, total communication round of FL \\ 
        $a$ & Attention level \\ 
        $N, N_{e, k, a}$ & FoV Tiles, tiles at attention level $a$ \\ 
        $r_{e, k, a}$ & Resolution ratio of attention level $a$\\ 
        $b_{e, k, a}$ & Resolution size of tile pixels for attention $a$  \\ 
        $b_{k, a, th}$ & Resolution threshold of tile pixels for attention $a$ \\ 
        $g_{e, k, a}$ & Tile size with attention level $a$\\ 
        $G_{e, k}$ & The GoP length\\ 
        $\text{QoE}_{e, k}(t)$ & QoE for user $k$ at $t$ \\ 
        $\text{hfQoE}_e(t)$ & Horizon-fair QoE over time horizon $t$\\ 
        $T_{e, k}^{(d)}, T_{e, k}^{(r)}$ & Time latency of downloading and rendering \\ 
        $T_{e, k}, T_{th}$ & Total time latency, time threshold \\ 
        $R_{e, k},  \Delta R_k$ & Theoretical transmission rate, estimated rate bias \\ 
        $B_{e, k}, B_{e, max}$ & Sub-channel bandwidth, total bandwidth \\ 
        $P_{e, k}$ & Transmit power from BS and $k^{t h}$ user \\ 
        $h_{e, k}$ & Rayleigh Channel gain \\ 
        $d_{e, k}$ & Distance between BS and the $k$-th user \\ 
        $I_{e, k}, \sigma_{e, k}^2$ &  Inter-cell interference,  noise power \\ 
        $\alpha, \omega$ & Path loss exponent, compression ratio \\ 
        $c_a$ & Number of cycles for processing a bit at level $a$ \\ 
        $f_{e, k}, \Delta f_k$ & Allocated CPU frequency, estimated rate bias \\ 
        $f_{e, max}$ &   Maximum CPU frequency \\ 
        $\mathcal{D}_e, \mathcal{D}$ & MEC local dataset, global dataset of the MEC system  \\
        $\mathbf{w}_e, \mathbf{w}$  & MEC local model, global model  of the MEC system \\
        $\boldsymbol{R}, \boldsymbol{\hat{R}}, \boldsymbol{S}, \boldsymbol{A}$ & Reward, reward-to-go, state, action  \\
        $\boldsymbol{U}$ & User numbers and levels information  \\
      $\tau$ & Trajectory consisting of  $(\boldsymbol{R}, \boldsymbol{S}, \boldsymbol{A})$\\
        $\tau^{(tr)},\tau^{(tr)}$ & Training and testing trajectory  \\
        $L_{tr},L_{te}$ & Training and testing trajectory length   \\
        $E, M$  & Local epoch,  local iteration in FL \\
        $\eta, B$  & Learning rate, batch size in FL \\
        \bottomrule 
    \end{tabular}
    }
    \vspace{-5pt}
\end{table}
 
\subsection{Contributions}

This paper proposes a FedPromptDT framework to address the resource allocation problem when the MEC  system provides customized  VR services for heterogeneous users.
Our contributions are summarized as follows:
\begin{itemize}
\item  We focus on enhancing the user's immersive experience in MEC-assisted VR services by exploring hierarchical attention levels based on human vision.
We introduce a QoE metric that integrates the MEC system's latency, user attention levels, and user-preferred resolutions to quantify individual user experience.


\item   We formulate a QoE-based maximization problem to enhance the user experience.
The problem aims to optimize resource allocation for CPU frequency, bandwidth, and user resolution while considering QoE and horizon-fair QoE constraints.  
We transform it into an RL problem to learn a generalized policy that applies to various user requirements across all MEC servers.

\item   We propose a novel FedPromptDT framework for learning the generalized policy.
This framework employs FL across MEC servers for pre-training a FedPromptDT model through prompt-based sequence modeling.
MEC servers can use prompts to aid pre-trained models in perceiving a user environment and generate optimal allocation without re-training or online turning.

\item    Our extensive experimental evaluations and ablation studies demonstrate that the pre-trained FedPromptDT model surpasses baseline methods and displays remarkable adaptability, maintaining superior performance across various user environments.
\end{itemize}

\subsection{Related Works}

\subsubsection{Customized VR resolution}
Despite the explosive expansion of the VR market, there are still large gaps between the huge demand for VR content and the infrastructure's capacity, particularly for VR streaming, also called 360° video streaming. One advancement in video streaming involves adaptive tile-based techniques, delivering VR content by dividing the 360° video into temporal segments and spatial tiles \cite{8717893}. As users typically focus on a restricted portion of the video, known as the viewport, each tile can be individually requested at varying quality levels, prioritizing content within the viewport. Attention-based mechanisms are guidelines for adjusting the tile quality level \cite{8824804,yu2023attention}. By employing attention, the system can dynamically allocate higher quality levels to specific tiles based on the user's focus or interest, such as the content awareness \cite{9018228,10144339} and visual-attention awareness \cite{8717893,8626197} within the viewport. Content-based attention might be more suitable when the goal is to deliver specific content elements or details that contribute to the overall understanding of the video. Differently, visual-based attention is beneficial when the primary objective is to enhance the viewer's experience by emphasizing high-quality rendering in areas where their eyes are directed, making it particularly effective for immersive experiences.

This paper emphasizes the immersive experience for VR users and explores three attention levels based on the hierarchical nature of the human eye. 
We also emphasize user preferences, recognizing that individuals may have varying priorities regarding their viewing experience. 
By offering a spectrum of resolution levels, our approach enables users to tailor their settings according to their preferences and the capabilities of their devices.
Therefore, this user-centric flexibility ensures a more customized VR experience.

\subsubsection{Federated learning for reinforcement learning}

FL \cite{mcmahan2017communication} has emerged as a promising paradigm for collaborative model training while preserving data privacy.
When intersected with RL, FL offers unique solutions to decentralized decision-making problems. 
The concept of federated RL was first explored by \cite{Saurabh2017FMADRL} to maintain data privacy and achieve greater learning efficiency in multi-agent distributed RL.
Since then, numerous studies have advanced the field by addressing key challenges such as heterogeneous environments \cite{JinPYWZ22FRL}, communication efficiency \cite{KhodadadianSJM22}, and algorithmic stability \cite{FanMDJTL21},  while implemented into different scenarios such as robotic system navigation \cite{LiuWL19FRL}, edge caching \cite{WangLWLTL21} and MEC resource management \cite{YuCZGW21}.
 These works provide a comprehensive foundation for integrating FL with RL, showcasing the potential to enable distributed and privacy-preserving in MEC systems.

This paper aims to optimize resource allocation for MEC servers to enhance users' QoE in the MEC system amidst diverse user environments.
Despite progress in federated RL, existing methods often produce sub-optimal policies that cannot generalize well to diverse environments \cite{li2023RLTransformers}.
When implemented in our task, these methods require re-training or online fine-tuning to adapt to various user environments during online execution.  
In contrast, our FedPromptDT framework can address such limitations by prompting the offline-trained model, allowing for rapidly adapting diverse user environments without re-training.

\subsubsection{Decision transformer for reinforcement learning}
DT, introduced by \cite{Chen2021decision}, is a significant paradigm shift in RL.
It moves away from traditional value function approximation and towards a model that transforms the RL problem as a sequence modeling task.
By leveraging the powerful capabilities of transformer architectures, originally popularized in natural language processing tasks, DT can directly model the relationship between states, actions, and returns.
 This eliminates the need for explicit policy or value function estimation and yields competitive results on a range of benchmark RL tasks \cite{Chen2021decision,xu2022PromptDT,li2023RLTransformers}. 
Since then, prompt-based methods have been integrated into DT to enhance its generalization capabilities in multiple tasks. 
These methods, such as text prompts in multi-modal household tasks \cite{ShridharTGBHMZF20}, trajectory prompts in multi-control tasks \cite{xu2022PromptDT}, and goal prompts in clinical recommender systems \cite{Seunghyun2023ClinicalDT}, provide task-specific instructions for DT to adapt to specific tasks without modifying the model parameters.

This paper introduces FedPromptDT to tackle the resource allocation problem of MEC servers under multiple user tasks.
The MEC servers focus on enhancing customized user experiences by achieving optimal resource allocation for various user environments.
Our prompt design incorporates environmental cues and user-preferred allocation to enable automatic prompting of the pre-trained model, eliminating the need for human annotation as required in text prompts \cite{ShridharTGBHMZF20} or sampling from human experts during execution as required in trajectory prompts \cite{xu2022PromptDT}.
Moreover, training PromptDT requires diverse trajectories to handle highly personalized user data, which may not be directly applicable to local training at each MEC server due to the limited data or centralized training at the cloud due to strict privacy requirements.
Therefore, we employ FL to pre-train a FedPromptDT model, which provides a privacy-preserving and scalable solution for customized VR services that cater to user preferences.

The remainder of this paper is organized as follows: Section \ref{sec:Preliminaries} presents the preliminaries, while Section \ref{sec:system} formulates the system model and problem. 
In Section \ref{sec:algo}, we propose our solution method, FedPromptDT, by transforming the problem.
The performance evaluation and ablation study on FedPromptDT are presented in Section \ref{sec:evaluation}. Finally, Section \ref{sec:conclusion} concludes our contribution and findings.
Besides, Table \ref{tab:notation_list} summarizes the primary notations used throughout the paper.

\section{Preliminaries}\label{sec:Preliminaries}
\subsection{Federated Learning on Distributed MEC Severs}
We consider an FL framework on a distributed MEC system with $E$ MEC servers, each possessing its own dataset $\mathcal{D}_e $ 
consisting of $n_e$ data samples.
A union of all MEC datasets refers to the global dataset of the MEC system, represented by $\mathcal{D} = \cup_{e=1}^E \mathcal{D}_e$
with a total of $n = \sum_{e=1}^E n_e$ data samples.

\subsubsection{Enhance MEC's model generalization with federated learning}
\begin{figure}[th]
    \centering
    \includegraphics[width=0.5\textwidth]{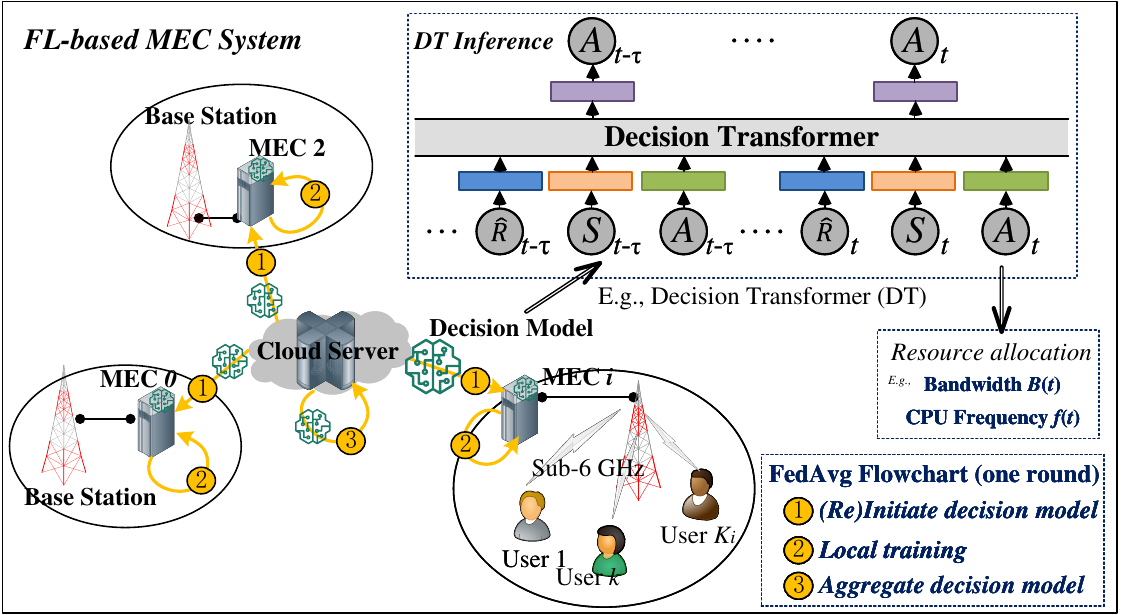}
    \caption{Illustration of the FL-based MEC system. 
    The system takes FL to enhance the generalization of its decision model on MEC servers.
   The decision model is considered a Decision Transformer (DT) that tokens states, actions, and returns of MEC servers using their corresponding linear embedding layers to predict actions for resource allocation.}
    \label{fig:FedDT}
\end{figure}

Different MEC servers meet diverse user requirements (e.g., time latency, preferred resolution, and so on) and user equipment (e.g., computation ability), leading to heterogeneous MEC data, denoted by $\mathcal{D}_e   \neq  \mathcal{D}_{e^\prime} \neq \mathcal{D}$ when $e \neq e^\prime$.
That is, a model trained on a single MEC may not be sufficiently equipped to satisfy the diverse needs of all users, while collecting all MEC data in the cloud may leak user privacy.
As per a recent FL survey \cite{shao2023survey}, current FL methods like model-sharing-based algorithm \cite{mcmahan2017communication, li2020federated, zhou2022fedfa, li2023fedcir, zhou2023understanding} can effectively handle heterogeneous data and improve model generalization, compared with solo training.
Therefore, we employ FL to train a model across all MEC servers to improve the model generalization while protecting privacy, as shown in Figure \ref{fig:FedDT}.
Please refer to section \ref{section: HeterogeneousVR} for further information regarding the problem addressed in this work.

\subsubsection{Objective function of federated learning and its solution}
 FL  minimizes the expected global loss $  \mathcal{L}(\mathbf{w}) := \mathbb{E}_{\xi \in \mathcal{D}}[  l (\mathbf{w};\xi)]$ on the global dataset $\mathcal{D}$, where $l(\mathbf{w})$ denotes the global loss function for model  $\mathbf{w}$, and $\xi$ denotes a data sample belonging to $\mathcal{D}$.
In this work, we consider  a canonical FL solution, FedAvg \cite{mcmahan2017communication}, and reformulate the FL objective on MECs as:
\begin{equation}
 \begin{aligned}
   \min_{\mathbf{w} \in \mathbb{R}}  \mathcal{L}(\mathbf{w} )  = &    \sum_{e=0}^{E-1} \frac{n_e}{n} 
 \mathcal{L}_e(\mathbf{w} )  
    =     \sum_{e=0}^{E-1} \frac{n_e}{n} \sum_{i=1}^{n_e}  l_e (\mathbf{w};\xi_i\in  \mathcal{D}_e),
    \end{aligned}
   \label{fl}
   \vspace{-1mm}
\end{equation}
where $\mathcal{L}_e(\cdot)$ and 
$l_e(\mathbf{\cdot})$
are the expected and estimated local loss of the $e$-th MEC on its local dataset $\mathcal{D}_e$, respectively, and $\xi_i$ denotes a data sample belonging to $\mathcal{D}_{e}$.
In the MEC system, FedAvg optimizes the objective (\ref{fl}) by periodically averaging the models locally updated by MECs, which follows the steps of each round:
\begin{enumerate}
    \item MECs update their local models  $\{\mathbf{w}_e\}_{e=0}^{E-1}$ independently by minimizing their losses $\{ \mathcal{L}_e (\mathbf{w}_e) \}_{e=0}^{E-1}$ on $\{ \mathcal{D}_e \}_{e=0}^{E-1}$,  and upload the updated models to the cloud server;
    \item The cloud server  aggregates  local models to obtain a new global model, denoted by $\mathbf{w} =\sum_{e=0}^{E-1} \frac{n_e}{n}\mathbf{w}_e$, and broadcasts the global model $\mathbf{w}$ to MEC servers;
    \item   MECs re-initialize their local models with $\mathbf{w}$ and perform local training of the next round.
\end{enumerate}
The aforementioned process continues until the global model reaches convergence.

\subsection{Decision Transformer}

 RL aims at learning a policy that maximizes the expected sum of rewards $\mathbb{E}\left[\sum_{t=1}^T r_t\right]$ along the whole Markov decision process in the environment space $(\mathcal{S}, \mathcal{A}, P, \mathcal{R})$.
The RL training process is based on states $\boldsymbol{S} \in \mathcal{S}$, actions $\boldsymbol{A} \in \mathcal{A}$, and a reward function $\boldsymbol{R} =\mathcal{R}(\boldsymbol{S}, \boldsymbol{A})$ without the need for the exact knowledge of transition dynamics $P\left(\boldsymbol{S}^{\prime} \mid \boldsymbol{S}, \boldsymbol{A}\right)$.
This work considers an RL framework that can only access some fixed limited datasets with offline training.
The dataset consists of some trajectories from arbitrary policies, where
one  trajectory is denoted by 
$\tau=\left(\boldsymbol{S}_0, \boldsymbol{A}_0, \boldsymbol{R}_0, \cdots,\boldsymbol{S}_t, \boldsymbol{A}_t, \boldsymbol{R}_t, \cdots, \boldsymbol{S}_T, \boldsymbol{A}_T, \boldsymbol{R}_T\right)$,
where the subscript $t$  refers to the trajectory timestep.

In this framework, the RL agent has limited ability to explore the environment and cannot obtain data through interactions with the environment.
This consideration is because the MEC system encounters diverse user environments that are agnostic to the MEC servers in the section \ref{sec:system}.
Then, this work takes offline training to pre-train a decision transformer model and online prompting to adapt to user environments in Section \ref{sec:algo}.
The model can generate optimal allocation without requiring re-training or online-turning the model.

 \subsubsection{Transformer architecture and Generative Pre-trained Transformer (GPT)}
 Vaswani et al. \cite{Vaswani2017transformer} first proposed the Transformer architecture to model sequential data efficiently. 
A normal transformer architecture uses one unit to encode the input, called the Encoder, and a separate unit to generate the output, called the Decoder. There are two types of attention during inference: self-attention and encoder-decoder attention.
Each attention layer takes $m$ embeddings $\left\{x_i\right\}_{i=1}^m$ as input tokens and outputs $m$ embeddings $\left\{z_i\right\}_{i=1}^m$, where the output dimensions are the same as the input. 
The $i$-th input token is transformed linearly into a key $k_i$, a query $q_i$, and a value $v_i$.
The $i$-th output of the self-attention layer is determined by weighting the values $v_j$ by the normalized dot product between the query $q_i$ and other keys $k_j$, which is formulated as follows:
\begin{equation}
  z_i=\sum_{j=1}^n \operatorname{softmax}\left(\left\{\left\langle q_i, k_{j^{\prime}}\right\rangle\right\}_{j^{\prime}=1}^n\right)_j \cdot v_j.  
  \label{eq:transformer}
\end{equation}
This allows the layer to implicitly associate different input tokens based on the similarity between the query and key vectors. 
In contrast, GPT \cite{radford2018improving_GPT1,Brown2020GPT3} is a decoder-only transformer with a single unit for encoding the input and generating the output with masked self-attention.
Taking (\ref{eq:transformer}) as an example, it modifies the Transformer architecture to enable autoregressive generation, using the masked self-attention to replace the softmax over the $n$ tokens with the previous $j$ tokens in the sequence $j \in[1, n]$.

\subsubsection{Decision transformer}
DT follows the rationale of the GPT architecture in autoregressive language modeling to abstract offline RL as a sequence modeling problem.
Specifically,  with autoregressive modeling, DT generates future actions by conditioning past states, actions, and desired rewards to make decisions, and its trajectory is formulated as follows: 
$$
\tau=\left(\hat{\boldsymbol{R}}_1, \boldsymbol{S}_1, \boldsymbol{A}_1, \cdots, \hat{\boldsymbol{R}}_t, \boldsymbol{S}_t, \boldsymbol{A}_t, \cdots, \hat{\boldsymbol{R}}_T, \boldsymbol{S}_T, \boldsymbol{A}_T\right),
$$
where the rewards-to-go 
$\hat{\boldsymbol{R}}_t=\sum_{t^{\prime}=t}^T r_{t^{\prime}}$
denotes the sum of future rewards from timestep $t$ to $T$.
The key difference to common RL methods is to replace the reward 
$\boldsymbol{R}_t$
in the trajectories as   
$\hat{\boldsymbol{R}}_t$.
This helps generate actions based on future desired returns rather than past rewards. 
As shown in Figure \ref{fig:FedDT}, DT leverages autoregressive sequence modeling to learn the pattern behind states, actions, and rewards rather than fitting value functions or computing policy gradients.

\section{System Model and Problem Formulation}
\label{sec:system}
\begin{figure}[th]
  \centering
  \subfigcapskip=-2pt 
  \hspace{-4mm}
    \subfigure[Attention-based VR content illustration for heterogeneous users.]
        {\includegraphics[width=3.6in]{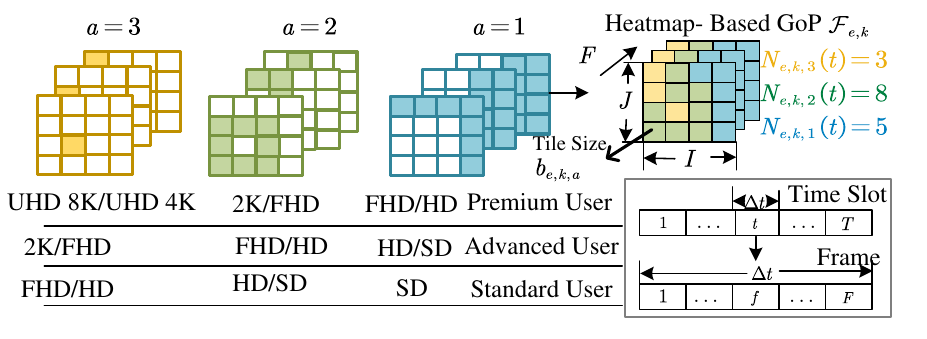}
        \label{fig:FOV} 
        }
    \hspace{-10mm}
    	\subfigure[Eye gaze location of each frame.]{
		\includegraphics[width=0.235\textwidth]{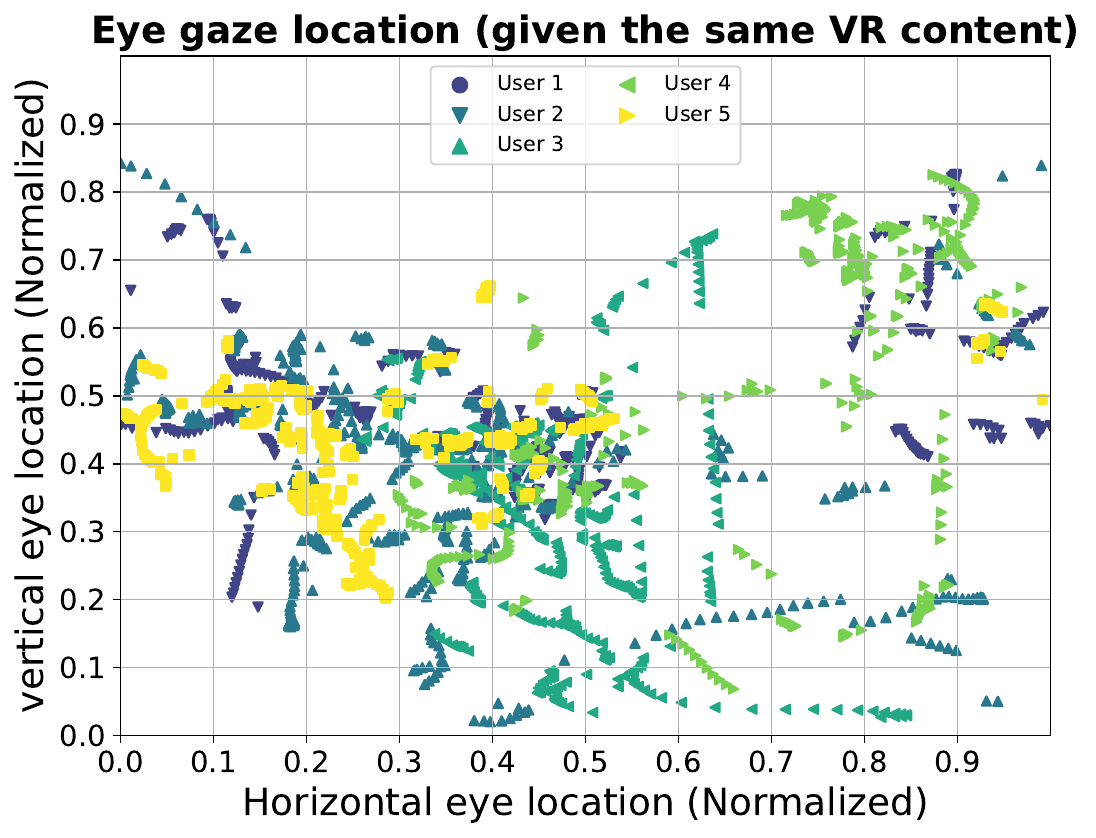}
        \label{fig:eye_gaze_location}}
    \hspace{-4mm}
	\subfigure[Data size with varying user level.]{
		\includegraphics[width=0.235\textwidth]{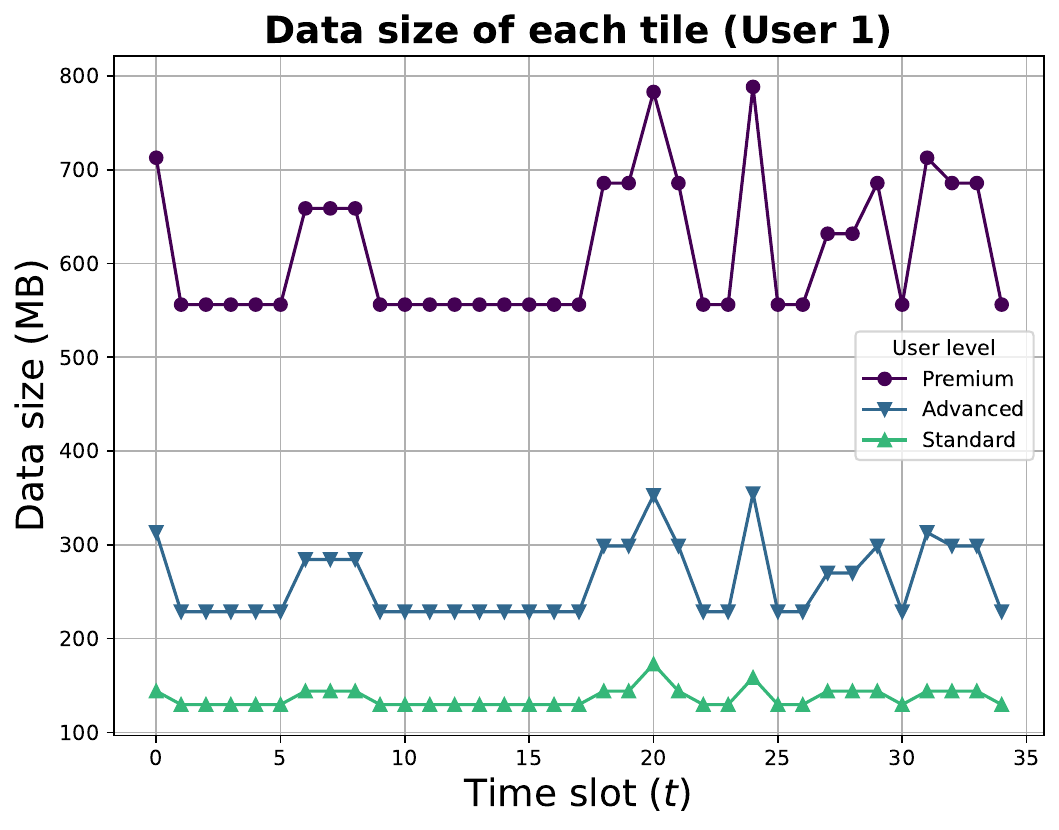}
        \label{fig:tile_size}}
    \caption{ (a) Illustration of the attention-based VR video content for the $k$-th user served by $e$-th edge server at time slot $t$. 
    (b) Different users have diverse eye gaze locations while viewing the same 360° VR video content from \cite{Xu2018video360}, indicating heterogeneous attention tile size.
    (c) Different user levels are required to process different tile sizes, resulting in heterogeneous computation and communication requests.}
  \label{fig:motivation}
\end{figure}



In this work, each MEC server renders a different group of pictures (GoPs) of VR content for heterogeneous user requests.
MEC servers have sufficient computing and storage resources to stream reliable VR content to different users.
These servers are placed near a base station (BS) that uses sub-6GHz technology for communications. Additionally, we consider the digital replica of the MEC systems\cite{yu2022bi}. 
These digital MECs serve the purpose of offering both historical data and real-time updates, enriching the overall analytical capabilities and decision-making processes within heterogeneous environments.

\subsection{Customized Resolution for Users' VR Content} 

Human vision hierarchy is structured around three primary levels: central, paracentral, and peripheral vision  \cite{8824804}.
The central vision offers the sharpest quality but spans less than 5\% of the visual field;
para-central vision perceives color and accounts for approximately 30\%; 
peripheral vision detects motion and constitutes around 60\%.
As depicted in Figure \ref{fig:FOV}, the 360° VR video content with spherical features is projected onto a two-dimensional (2D) plane.
For the $k$-th user ($k\in \mathcal{K}_e=[0,1,\cdots,K_e)$) served by the $e$-th MEC server ($e\in\mathcal{E}=[0,1,\cdots,E)$) at time slot $t$, the video within the FoV $\mathcal{F}_{e,k}(t)$ is uniformly cropped into $I\times J=N$ tiles. 

Inspired by the hierarchical human vision, we define the visual attention level ($a\in \{1,2,3\}$) and assign the tiles to three resolution levels accordingly. 
At time slot $t$, each attention level has $N_{e,k,a}(t)$ tiles and remains fixed for a group of frames $F$, referred to as a GoP $G_{e,k}(t)=\sum_{a}g_{e,k,a}(t)$. 
The data size of each tile at attention level $a$ is $g_{e,k,a}(t)=N_{e,k,a}(t)\times b_{e,k,a}(t) \times F$, where $b_{e,k,a}(t)$ is the size of each tile at attention level $a$ and with a resolution $r_{e,k,a}(t)$ at time slot $t$. 
The higher the attention level $a$ and the resolution $r$, the higher size of $b_{e,k,a}(t)=r_{e,k,a}(t)\times b_{max}$. 
Due to the requirements of user-preferred resolutions, we define three different user levels as follows:
\begin{itemize}
    \item Premium users: The three attention levels correspond to three resolutions when encoding the video frames.
    Premium users enjoy the highest quality and immersive attention-based visual experience with ultra-high definition (UHD) 8K resolution ($7680 \times 4320$) or 4K resolution ($3840 \times 2160$ pixels) at attention level $a=3$, followed by 2K resolution ($2048 \times 1080$ pixels) or full high definition (FHD) ($1920\times 1080$ pixels) at attention level $a=2$, and FHD or high definition (HD) resolution ($1280 \times 720$ pixels) at attention level $a=1$.
    \item Advanced users: The three attention levels cater to advanced users with 2K or FHD at attention level $a=3$, FHD or HD at attention level $a=2$, HD or standard definition (SD) resolution ($640 \times 480$ pixels) at attention level $a=1$. 
    \item Standard users: The three attention levels are designed for entry-level users to view video frames at FHD resolution or HD at $a=3$, HD resolution or SD at $a=2$, and SD at $a=1$.
\end{itemize}

\subsection{Measuring Quality of Experience for Heterogeneous Users}\label{section: HeterogeneousVR}

Different users have varying preferences for VR content, e.g., diverse gaze attention and resolution levels on the same content, as shown in Figures \ref{fig:eye_gaze_location} and \ref{fig:tile_size}.
This results in the MEC servers needing to render different tiles and data sizes.
Moreover,   different MEC servers need to dynamically serve different numbers of users due to user mobility.
Thus, optimizing resource allocation for MEC servers across diverse user environments, including user preferences and user numbers, is important.

Based on the Weber-Fechner Law\cite{Reichl2010, 10144339}, we introduce a novel concept called attention-based resolutions, which assists in quantitatively evaluating the QoE for individual users: 
\begin{align}
        \text{QoE}_{e,k}(t)=  \left(1-\frac{T_{e,k}}{T_{th}} \right) \sum_{a} \frac{a N_{e,k,a} (t)}{N} \text{ln}\left(1+\frac{ b_{e,k,a}(t) }{b_{k,a,th}} \right),
    \label{qoe}
\end{align}
where $T_{e,k}$ is the total latency, $T_{th}$ is the maximal threshold of the latency, $b_{k,a,th}$ is the smallest tile resolution for different user level on each attention level $a\in \{1,2,3\}$.
If the user $\text{QoE}_{e,k}(t)$ is higher than a predefined $\text{QoE}_{e, th}$,  the user's experience surpasses their minimum expectations; vice verse.
According to (\ref{qoe}), the QoE metric depends on the system latency, resolution requirements, user level, and tile sizes. 
That is, when user preferences and user numbers are heterogeneous,   various user tile sizes $N_{e,k,a}$ and resolutions $b_{k,a,th}$  would induce various  $\text{QoE}_{e,k}$ across inter-MEC and intra-MEC.
Consequently, the MEC server should uniquely allocate communication and computation resources to each user to improve their VR experience.

\subsection{Digital Twin-empowered MEC}
The digital twin-empowered MEC system\cite{yu2022bi} is an emerging architecture that combines MEC with digital twin technologies. 
This hybrid architecture monitors both computational aspects (e.g., the CPU's clock frequency) and communication elements (e.g.,  bandwidth and the state of the edge network).
This enables real-time visibility into the state of MEC systems, facilitating the collection of the history data and real-time updates for various environments.
By harnessing these datasets, the MEC system can effectively train its decision-making models to identify the most effective resource distribution strategies.
Consequently, its MEC servers make informed decisions and provide recommendations to guide users' optimal assignment of physical resources.

The digital user is the digital replica of the physical user's features. 
To assist  MEC servers with making resource allocation decisions, 
the digital user at time slot $t$ is defined as
\begin{align}
    \boldsymbol{D}_{e,k}(t)=\{N_{e,k,a}(t),\text{QoE}_{e,k}(t)\}, a\in \{1,2,3\},
    \label{digital_user}
\end{align}
where the tiles number at each attention level  $N_{e,k,a}(t)$ of the user's FoV  $\mathcal{F}_{e,k}(t)$ is well-collected and predicted in the  digital user. 
The digital replicas are heterogeneous since user preferences and numbers vary.

The digital MEC is the digital representation of a MEC server. 
We define the $e$-th digital MEC at time slot $t$ as:
\begin{align}
   \boldsymbol{D}_{e}(t)= \{ \boldsymbol{D}_{e,k}(t),  R_{e,k}(t),  f_{e,k}(t), 
   \text{hfQoE}_e(t) ; k\in \mathcal{K}_e\},
\end{align}
where $\boldsymbol{D}_{e,k}(t)$ is digital users defined in (\ref{digital_user}), $R_{e,k}(t)$ is the communication transmission rate of the $k$-th user,  $f_{e,k}(t)$ is the computation capacity assigned to the $k$-th user, $\text{hfQoE}_{e}(t)$  is the system fairness  of all users, and $K_e$ is the maximal user number accommodated by the $e$-th MEC.
Due to the heterogeneous digital users $\boldsymbol{D}_{e,k}(t)$ and their varying communication and computation requests, the digital replicas exhibit diversity across all MEC servers.

\subsubsection{Communication model}
For the sub-6 GHz link between BS and $k$-th user's head-mounted display, the theoretical transmission rate in the digital MEC is given as 
\begin{align}
R_{e,k}(t)=B_{e,k}(t)log_2\left(1+\frac{P_{e,k}(t)h_{e,k}(t)\left(d_{e,k}(t)\right)^{-\alpha}}{I_{e,k}(t)+\sigma_{e,k}^{2}}\right),
\end{align}
where $B_{e,k}(t)$ and $P_{e,k}(t)$ are the sub-channel bandwidth and the transmit power of the BS to $k$-th user at time slot $t$,
$h_{e,k}(t)$ is the Rayleigh channel gain, 
$d_{e,k}(t)$ is the distance between BS and $k$-th user,
$\alpha$ is the path loss exponent, 
$I_{e,k}(t)$ denotes the inter-cell interference,
and $\sigma_{e,k}^{2}$ is the noise power of the sub-6 GHz link\cite{9536410}.

The calibrated communication latency from the edge BS to the $k$-th user over wireless links is computed as
\begin{align}
T_{e,k}^{(d)}(t)=\frac{G_{e,k}(t)}{\omega \left(R_{e,k}(t)-\varDelta R_{e,k}(t)\right)},
\end{align}
where $\omega$ is the compression ratio before transmission, $\varDelta R_{e,k}(t)$ is the estimated rate bias between the theoretical transmission rate in the digital MEC and the actual transmission rate retrieved from the feedback in the physical world.

\subsubsection{FoV rendering model}
The resource allocation for the $k$-th user computing is expressed as ${f}_{e,k}(t)$, indicating the allocated computing capacity for rendering GoP $G_{e,k}(t)$ at time slot $t$.
Given $f_{max}$ denoting the server's maximum computing capacity, the calibrated rendering latency of $k$-th user's requested FoV is computed as
\begin{align}
T_{e,k}^{(r)}(t)= \frac{\sum_a{{g_{e,k,a}(t)}c_{a}}}{f_{e,k}(t)-\varDelta f_{e,k}(t)},
\end{align}
where $\varDelta f_{e,k}(t)$ is the estimated CPU frequency bias\cite{9174795} between the estimated CPU frequency in the digital MEC and the actual CPU frequency, retrieved from the physical world, $c_{a}$ denotes the number of cycles required for processing one bit of input data at attention level $a$. 
The total latency for displaying the requested FoV by the $k$-th user, including rendering and downlink latency, is calculated as:
\begin{align}
    T_{e,k}(t)=T_{e,k}^{(d)}(t)+T_{e,k}^{(r)}(t).
    \label{time_latency}
\end{align}
\subsubsection{Horizon-fair QoE}
Horizon-fair QoE \cite{si2022enabling,7588099} over time horizon $t$ is computed as
\begin{align}
    & \text{hfQoE}_e(t)  =1-\frac{\sigma^{(\text{hfQoE})}_e}{\sqrt{K_e}},
\end{align}
where $\sigma^{(\text{hfQoE})}_e$ is the standard deviation of the users' average QoE and formulated as:
\begin{align}
\sigma^{(\text{hfQoE})}_e=\sqrt{\frac{1}{K_e}\sum_{k=1}^{K_e}\left(\text{avgQoE}_{e,k}(t)-\overline{\text{avgQoE}_{e}(t)}\right)^2}.
\end{align}
Here,   $\text{avgQoE}_{e,k}(t)=1/t \left(\sum_{t=1}^{t}\text{QoE}_{e,k}(t)\right)$  is the average QoE of the $k$-th user over time horizon $t$, 
and $\overline{\text{avgQoE}_{e}(t)}$ is the average of all $K$ users' $\text{avgQoE}_{e,k}(t)$ at the time slot $t$.

\subsection{Problem Formulation}
The resource allocation behavior in the digital MEC is formulated to maximize the long-term $\text{QoE}$ for immersive VR experience by jointly optimizing the attention level-based tile resolution ratio $\boldsymbol{r}=\{r_{e,k,a}(t)\},k\in \mathcal{K}, a\in \{1,2,3\}$, bandwidth $\boldsymbol{B}=\{B_{e,k}(t),k\in \mathcal{K}_{e}\}, e\in \mathcal{E}$,  and assigned CPU frequency $\boldsymbol{f}=\{f_{e,k}(t),k\in \mathcal{K}_{e}, e\in \mathcal{E}\}$ in $T$ time steps. 
Then, the problem is formulated as
\begin{subequations}
\begin{align}
    (\textbf{P}_0) \quad
    &\max_{\boldsymbol{r},\boldsymbol{B},\boldsymbol{f}}\quad \sum_{t=0}^{T}\sum_{e=1}^{E}\sum_{k=1}^{K_e}\text{QoE}_{e,k}(t)
    \label{a} \\
    \textbf{s.t.} \quad 
    &\sum_{k} B_{e,k}(t)\leqslant 
    B_{e,max},\quad \forall k\in \mathcal{K}_e, \forall e\in \mathcal{E},
    \label{b}\\
 &\sum_{k} f_{e,k}(t)\leqslant f_{e,max},\quad \forall k\in \mathcal{K}_e,\forall e\in \mathcal{E},
    \label{c}\\
    & \text{QoE}_{e,k}(t)\geqslant \text{QoE}_{e,k,th},\quad \forall k\in \mathcal{K}_e,\forall e\in \mathcal{E},
    \label{d}\\
    & \text{hfQoE}_e(t)\geqslant \text{hfQoE}_{e,th},  \quad \forall e\in \mathcal{E}.
    \label{e}
\end{align}
\end{subequations}
The constraint (\ref{b}) denotes that the system bandwidth cannot exceed the total bandwidth $B_{e,max}$ of the $e$-th MEC server. 
The constraint (\ref{c}) denotes that the system CPU frequency cannot exceed the maximum frequency $f_{e,max}$ of the $e$-th MEC server.
The constraint (\ref{d}) satisfies the long-term QoE fairness of the system at the $e$-th MEC server.

The optimization challenge presented in (\ref{a}) is characterized by long-term stochastic dynamics, encompassing several adaptive decision variables—specifically, $\boldsymbol{b}$, $\boldsymbol{B}$, and $\boldsymbol{f}$—within a dynamic system. 
Conventional optimization methods, like convex optimization, encounter notable difficulties when attempting to quickly find optimal solutions in high-dimensional spaces. 
In contrast, RL emerges as a promising strategy for addressing the problems featuring expansive action spaces.

\section{Proposed Method}
\label{sec:algo}
This section will discuss how to transform the problem ($\textbf{P}_0$) into an RL problem.
We then propose an FL and prompt-based DT method to address various user environments without retraining the model obtained by offline training. 
The method accommodates different user numbers and user preferences (i.e., user levels) across all MEC servers.

\subsection{Problem Transformation based on RL}
 \subsubsection{Our RL framework}
When solving problem ($\textbf{P}_0$) with the RL framework at the $e$-th MEC server, we formulate the state space $\mathcal{S}_e$, action space $\mathcal{A}_{e}$, and reward function $\mathcal{R}_{e}$ for $\forall e \in [0, E)$ as follows:
\begin{itemize}
    \item \textbf{State space}: The state $\boldsymbol{S}_e(t) \in \mathcal{S}_e$ at each time slot $t$ is represented as 
    \begin{align}
    \begin{split}
            \boldsymbol{S}_e(t)=&\left\{\boldsymbol{D}_{e}(t-1),\boldsymbol{D}_{e}(t), T_{e,k}^{(d)}(t), T_{e,k}^{(r)}(t)\right.,\\&\left.T_{e,k}(t); k\in \mathcal{K}_e\right\},
    \end{split}
    \label{eq:state}
    \end{align}
    with the information acquired from the digital users and the $e$-th digital MEC.
    \item \textbf{Action space}: The action $\boldsymbol{A}_{e}(t) \in \mathcal{A}_{e}$ of the local $e$-th MEC server at each time slot $t$ is formulated as follows:
    \begin{equation}
        \boldsymbol{A}_{e}(t)=\left\{\boldsymbol{A}_{e, k}(t); k \in\mathcal{K}_e \right\} ,
    \end{equation}
    with the resource allocation decision $\boldsymbol{A}_{e, k}(t)= \{{r}_{e, k,a}(t),{B}_{e, k}(t),{f}_{e, k}(t); a \in \{1,2,3\}\}$ on the $k$-th user, where  ${r}_{e, k, a}$ is the resolution decision of the attention level $a$, ${B}_{e, k}$ is the bandwidth decision, ${f}_{e, k}$ is the frequency decision. 
    \item \textbf{Reward function}: The reward function of the $e$-th MEC server is designed based on its users' QoE as follows:
\begin{align}
    \begin{split}
    \boldsymbol{R}_{e}(t)
    =\sum_{k=1}^{K}\text{QoE}_{e,k}(t)-\varpi_{1}\sum_{k=1}^{K}q_{e,k}^{\text{QoE}}-\varpi_{2}q_e^{\text{hfQoE}},
    \label{rewardfunction}
        \end{split}
\end{align}
where $ \boldsymbol{R}_{e} \in \mathcal{R}_e$, $\varpi_{1}$ and $\varpi_{2}$ are the penalty coefficients with  $\varpi_{1}:\varpi_{2} = 1:K_e$.
The two penalty terms are expressed in the following formulations:
\begin{align}
    q_{e,k}^{\text{QoE}}=\begin{cases} 0,\quad  \text{QoE}_{e,k}(t)\geqslant \text{QoE}_{e,k,th}, \quad k \in\mathcal{K}_e,\\
\text{QoE}_{e,k,th},\quad \text{otherwise},\\
    \end{cases}
\end{align}
and
\begin{align}
    q_e^{\text{hfQoE}}=\begin{cases} 0,\quad  \text{hfQoE}_{e}(t)\geqslant \text{hfQoE}_{th}\\
\text{hfQoE},\quad \text{otherwise}.\\
    \end{cases}
\end{align}
\end{itemize}

\subsubsection{FL-based RL policy}
Offline RL can commonly succeed in episodic environments with non-mutational distributions to learn the optimal policy with limited trained data.
However, as discussed in Section \ref{section: HeterogeneousVR}, various digital users and digital MEC lead to heterogeneous environments across MEC servers.
This indicates that the MEC's policy obtained from local training on each MEC server may not be suitable for all environments with varying user numbers and levels.
Therefore, we use FL to learn a global policy $\chi$ to solve the problem ($\textbf{P}_0$).

Specifically, for all the $e \in [0,E)$, $\chi$ is a mapping from states $\boldsymbol{S}_e$ to the action $\boldsymbol{A}_e$ to maximize the accumulated rewards $\boldsymbol{R}_{e}(t)$  from $t \in [0,T)$ within a single episode. 
To guide the learning process, we adopt an episode-based objective as follows:
\begin{align}
 \max   \mathcal{J}_{e,ep}(\chi)=\mathbb{E}\left[\sum_{\tau=0}^{T-1}\sum_{k=0}^{K_e -1}\gamma^{\tau}\boldsymbol{R}_{e}(t)|\boldsymbol{S}_{e}(t)\right], \forall e \in [0,E)
    \label{global_objective}
\end{align}
where $\gamma\in [0,1)$ is the discount factor determining the weight of the future long-term reward, and $\gamma=0$ indicates that only the current time slot $t$ is considered.

\begin{figure*}[t]
  \centering
  \includegraphics[width=\textwidth]{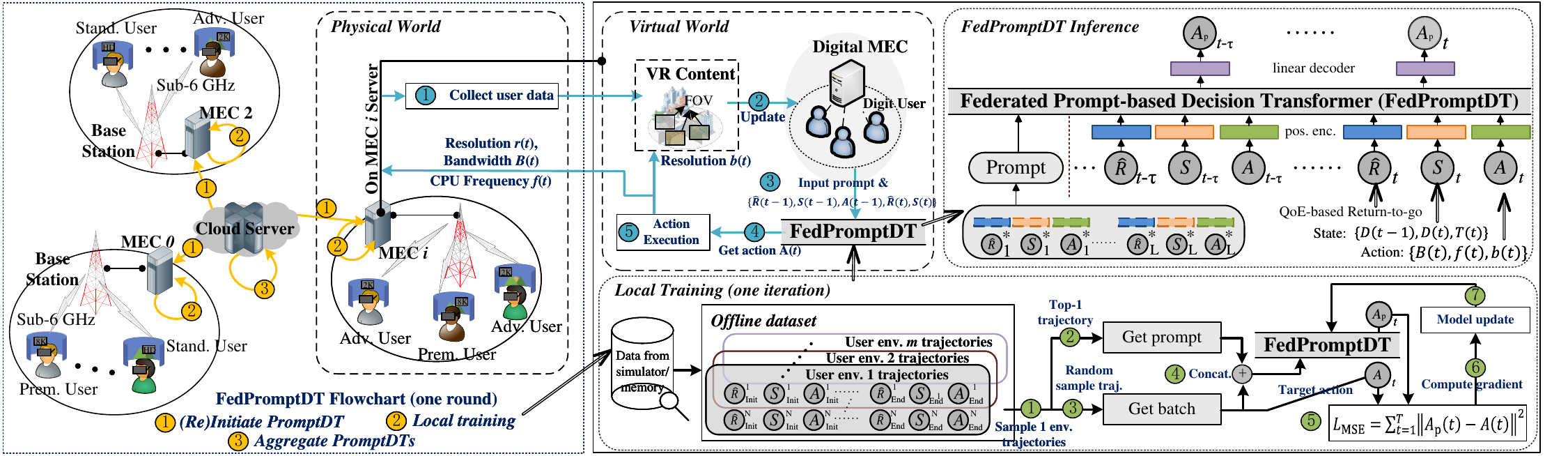}\\
  \caption{Illustration of the FedPromptDT-empowered MEC system. 
Digital twin allows the MEC system to monitor the system's real-time state, facilitate perceiving user environments, and collect historical data.
During online execution, states, actions, and returns are tokenized by their corresponding linear embedding layers and added with episodic timestep encoding.
These tokens are fed into the pre-trained FedPromptDT model to autoregressively predict actions with the prompt based on current user environments.
During local training in FL, the MEC server collects a batch that concatenates the prompt and training trajectories in each local iteration. After that, it updates FedpromptDT iteratively on various user environments throughout the local training process.
More details for offline training and online execution are in Section \ref{sec:OfflineFL_online_ex}.}
  \label{fig:systemmodel}
\end{figure*}
\subsection{Our Proposed Framework: FedPromptDT}
The meaning of (\ref{global_objective}) is to learn a generalized policy $\chi$  applicable across various user environments for all MEC servers.
The policy $\chi$  requires each MEC server to support varying user levels and numbers with optimal resource allocation.
These varying user statuses result in many potential user environments for RL-based algorithms to explore.
Common actor-critic algorithms such as Deep Deterministic Policy Gradient (DDPG) may fail to learn the optimal policy when the exploration is not efficient enough \cite{Matheron0S20Understanding}, i.e., the number of user environments explored is not enough in our RL tasks.
In contrast, transformer-based algorithms like DT  have stronger generalization potential to make allocation decisions on unseen user environments for MEC servers.



Inspired by the DT model, we propose FedPromptDT to learn a prompt-based generalized policy based on data stored in digital MEC with limited user environments.
The policy can be generalized to diverse user environments based on their corresponding prompts.
We illustrate the whole system flowchart and FedPromptDT architecture in Figure \ref{fig:systemmodel}, and summarize its FL-based offline training and online execution in Algorithms \ref{alg:FedPromptDT}, \ref{alg:Localtraining} and \ref{alg:PromptDT_Eval}.
Note that superscript $\cdot^{\star}$ is taken to distinguish the trajectories for prompting from the trajectories for training, and subscript $\cdot_{e}$ of the $e$-th MEC is omitted for simplicity in the following subsection.

\subsubsection{FedPromptDT model architecture}

The entire architecture of FedPromptDT is illustrated in Figure \ref{fig:systemmodel}.
The main architecture follows the GPT structure and involves three trainable linear layers to embed the tokens of reward-to-go, state, and action. 
The input dimensions of these linear layers, i.e., the dimensions of $\boldsymbol{S}$ and $\boldsymbol{A}$ tokens, are inconsistent since user environments have varying user numbers.
Therefore, we consider a maximal user number $K_{max}$ for all MEC servers to support consistent dimensions of $\boldsymbol{S}$ and $\boldsymbol{A}$ tokens.
Specifically, when $K_{max}$ is greater than the maximal user number $K_{max}^{(m)}$ of the $m$-th user environment,  the all elements related to the $k$-th user in the vector $\boldsymbol{S}$ and vector $\boldsymbol{A}$  are padded as zeros, $\forall k \in [K_{max}^{(m)},K_{max}), k\in \mathbb{N}$.
Furthermore, we use a trainable linear layer to add the same positional embedding to reward-to-go, state, and action embeddings corresponding to the same timestep in the trajectory $\tau$.   

Similar to the token embedding for the training trajectory, the prompt trajectory is also tokenized by three trainable linear layers, and positional embedding is added by a trainable linear layer.
 When considering a training trajectory $\tau$ with length $L_{tr}$ and a prompt trajectory  $\tau^{\star}$ with length $L_{pr}$, FedPromptDT takes $\tau^{\text {input }}=\left(\tau_i^{\star}, \tau_i\right)$ as sequential input.
 The input sequence corresponds to $3\left(L_{tr}+L_{pr}\right)$ tokens in the standard Transformer model. 
 Meanwhile,  FedPromptDT autoregressively predicts $L_{tr}+L_{pr}$ action tokens by its output head.

The output head employs a trainable linear layer with \textit{sigmoid} activation to predict actions.
The action predicted by the head corresponds to state tokens in the input sequence;
i.e., the head predicts the action token when the reward-to-go and state tokens input the embedding layer.
The predicted action $\boldsymbol{\hat{A}}_{e}(t)$ for the $e$-th MEC server at the $t$-th step is represented as:
\begin{equation*}
    \boldsymbol{\hat{A}}_{e}(t)= \{\hat{r}_{e, k, a}(t), \hat{B}_{e, k}(t),\hat{f}_{e, k}(t); a\in \{1,2,3\},k \in \mathcal{K}_{max}\},
\end{equation*}
where $\mathcal{K}_{max}=\{1,\cdots,{K}_{e},\cdots,{K}_{max}\}$.
To clarify,  $\hat{r}_{e, k, a}(t)$, $\hat{B}_{e, k}(t)$, and $\hat{f}_{e, k}(t)$ are allocation ratios ranging from $[0,1]$ due to the \textit{sigmoid} activation, rather than the actual resource allocation values.
The MEC server assign the $k$-th user attention-based resolutions by $b_{k, a, th}\times\hat{r}_{e, k, a}(t)$, allocate communication bandwidth by $B_{e, max}\times\hat{B}_{e, k}(t)/\sum_{k^\prime=1}^{K_e}\hat{B}_{e, k}(t)$ and decide computation frequency  by $f_{e, max}\times\hat{f}_{e, k}(t)/\sum_{k^\prime=1}^{K_e}\hat{f}_{e, k}(t)$.
Besides, the whole FedPromptDT model is optimized by minimizing the mean-squared error (MSE) loss $\mathcal{L}_{MSE}$ between predicted actions and corresponding ground-truth actions.

\subsubsection{Prompt design}

Prompts are crucial in guiding the Transformer models to perform specific tasks and generate responses. 
For instance,  text prompts are commonly employed to instruct the model on the desired output in natural language processing tasks \cite{SchickS21SmallLanguage}. 
Trajectory prompts are used in RL tasks to guide the model's actions \cite{xu2022PromptDT}. 
In this work, we design trajectory prompts to guide the FedPromptDT model to generate the desired actions in various user environments.

The prompt is designed as a segment of the top-1 trajectory when training FedPromptDT on a specific user environment of the data stored in digital MECs, as summarized in Algorithm \ref{alg:Localtraining}. 
Specifically, for the $m$-th user environment, the prompt trajectory  $\tau_m^{\star} =(\tau_{m,i}^{\star},\tau_{m,i+1}^{\star},\cdots,\tau_{m,i+L_{pr}}^{\star})$ is sampled from the top-1 trajectory based on episode (EP) rewards.
$\tau_{m,i}^{\star}=(\hat{\boldsymbol{R}}_i^{\star}, \boldsymbol{S}_i^{\star}, \boldsymbol{A}_i^{\star})$ consisting of the $i$-th reward-to-go $\hat{\boldsymbol{R}}_i^{\star}$, state $\boldsymbol{S}_i^{\star}$, and action $\boldsymbol{A}_i^{\star}$ in the top-1 trajectory.
The top-1 trajectory prompt is because it showcases the optimal resource decision path, which prompts the model to learn how to generate the optimal path during offline training.

The prompt can specify a user environment for the FedPromptDT model by implicitly demonstrating the transition dynamics $P(\boldsymbol{S}^{\star}_{i+1} \mid \boldsymbol{S}^{\star}_i, \boldsymbol{A}^{\star}_i)$ and its corresponding reward $\boldsymbol{R}^{\star}_i$ on the Markov decision process.
Meanwhile, the prompt length $L_{pr}$ is much smaller than the horizon of the whole trajectory.
The short prompt allows FedPromptDT to recognize the user environment while avoiding overfitting the environment of the top-1 trajectories, thus guiding it to learn a generalized policy.
On the other hand,  we augment the digital-MEC data by concatenating users' information $\boldsymbol{U}_m$  with the original states $\boldsymbol{S}_m(t)$ in (\ref{eq:state}) to explicitly prompt the information of user environments into the FedPromptDT model.
The concatenated information is formulated as follows:
\begin{equation}
    \hat{\boldsymbol{S}}_m(t)=(\boldsymbol{S}_m(t), \boldsymbol{U}_m),
    \label{eq:state_concate}
\end{equation}
where $\boldsymbol{U}_m=\{{U}_{m,0}, \cdots, {U}_{m,K_{max}^{(m)}}, \cdots, {U}_{m,K_{max}})\}$ incorporates the information of user numbers and levels.
We take hard coding for premium, advanced and standard levels as ${U}_{m,k}^{(pre)}=0.6$, ${U}_{m,k}^{(adv)}= 0.4$ and ${U}_{m,k}^{(sta)}=0.2$, respectively, and  ${U}_{m,k}=0$ for all $k \in [K_{max}^{(m)},K_{max})$
when $K_{max}>K_{max}^{(m)}$.
Thus, MEC servers can utilize the above implicit and explicit prompts to help  FedPromptDT perceive a user environment and generate optimal allocation actions.

\begin{algorithm}[t]
   \caption{FedAvg on PromptDT (FedPromptDT)}
   \label{alg:FedPromptDT}
\begin{algorithmic}
   \STATE {\bfseries  Input:} initial FedPromptDT model $\mathbf{w}^{(0)}$,  total MEC number \textit{E}, training round \textit{R},    data stored in digital MECs $\{\mathcal{D}_e\}_{e=0}^{E-1}$
   \FOR{each round $r= 0, \cdots, R-1$  }
       \STATE  Cloud server sends $\mathbf{w}^{(r)}$ to all MECs
       \STATE \textbf{on MEC }$e\in [0,E)$ \textbf{in parallel do}
            \STATE \quad \quad Initialize local FedPromptDT model
            $\mathbf{w}_e \gets \mathbf{w}^{(r)}$  
            \STATE \quad \quad Get the updated $\mathbf{w}_e^{(r)}=\textit{LocalTraining}(\mathbf{w}_e,\mathcal{D}_e)$  
            \STATE \quad \quad Send $\mathbf{w}_e^{(r)} $  back to the cloud server
        \STATE   \textbf{end on client}
        \STATE Cloud server gets the next-round global FedPromptDT by
        \STATE  $\mathbf{w}^{(r+1)} \gets   \sum_{e=0}^{E-1} \frac{n_e}{n}\mathbf{w}_e^{(r)}$ 
   \ENDFOR
   \STATE {\bfseries  Return:} Pre-trained global FedPromptDT model $\mathbf{w}^{(R)}$
\end{algorithmic}
\end{algorithm}

\begin{algorithm}[t]
   \caption{\textit{LocalTraining} on FedPromptDT}
   \label{alg:Localtraining}
\begin{algorithmic}
   \STATE {\bfseries  Initiate:} Local iteration $M$, batch size $B$, learning rate $\eta$, training trajectory length $L_{tr}$,  prompt trajectory length $L_{pr}$
   \STATE {\bfseries  Input:} FedPromptDT model $\mathbf{w}_e$,  digital MEC data $\mathcal{D}_e$
   \FOR{each iteration $m=0, 1, \cdots, M-1$}
        \STATE Sample data of a user environment  $\mathcal{D}_{e,m} \subset \mathcal{D}_e$ 
        \STATE Augment user information $\boldsymbol{U}_m$ into  $\mathcal{D}_{e,m}$
        \STATE Sample the top-1 trajectory of length $L_{pr}$ from $\mathcal{D}_{e,m}$ as 
        \STATE prompt $\tau_{e, m}^{(tr)\star} $ 
        \FOR{$b=0, 1, \cdots, B-1$}
            \STATE Sample a trajectory $\tau_{e, m, b}$ of length $L_{tr}$ from $\mathcal{D}_{e,m}$
            \STATE Concatenate $(\tau_{e, m}^{(tr)\star}, \tau_{e, m, b})$ as model input $\tau_{e, m, b}^{\text {(input)}}$
       \ENDFOR
       \STATE Get a minibatch $\mathcal{B}_{e, m}=\left\{\tau_{e, m, b}^{\text {(input) }}\right\}_{b=0}^{B-1}$  
        \STATE Get predicted action $\hat{\boldsymbol{A}}_{e}   = \mathbf{w}_e^{(m)}\left(\tau^{\text {input }}\right), \forall \tau^{\text {input }} \in \mathcal{B}_{e, m}$
        \STATE Compute $\mathcal{L}_{M S E}=\frac{1}{|\mathcal{B}_{e, m}|} \sum_{\tau \in \mathcal{B}_{e, m}}\left({\boldsymbol{A}}_{e}-\hat{\boldsymbol{A}}_{e}\right)^2$
        \STATE $\mathbf{w}_e^{(m+1)} \leftarrow \mathbf{w}_e^{(m)} - \eta \nabla_{\mathbf{w}_e^{(m)}} \mathcal{L}_{M S E}$
   \ENDFOR
   \STATE {\bfseries  Return:} Local FedPromptDT model $\mathbf{w}_e^{(M)}$
\end{algorithmic}
\end{algorithm}

\begin{algorithm}[t]
   \caption{FedPromptDT Inference on Execution}
   \label{alg:PromptDT_Eval}
\begin{algorithmic}
\STATE {\bfseries  Initiate:}  Test env. $\boldsymbol{S}(0)$, episode len. $T_{te}$, prompt len. $L_{pr}$
\STATE {\bfseries  Input:}  FedPromptDT model $\mathbf{w}$,  target reward-to-go $\hat{R}^{\star}$
\STATE Get initial reward-to-go  $\hat{\boldsymbol{R}}(0) = \hat{R}^{\star}$ 
\STATE Get initial test trajectory $\tau=(\hat{\boldsymbol{R}}(0),\boldsymbol{S}(0))$
\STATE Construct execution prompt $\tau^{(te)\star}= (\hat{R}^{\star},\boldsymbol{S}(0),\boldsymbol{A}^{\star})*L_{pr}$
\FOR{$t = 0, 1, \cdots, T_{te}-1$}
    \STATE Concatenate $\tau^{\star}$ with $\tau$ as input $\tau_0^{\text {(input)}}=\left(\tau^{(te)\star}, \tau\right)$
    \STATE Get predicted action $\hat{\boldsymbol{A}}(t) =\mathbf{w}(\tau_{t}^{\text {input}})[-1]$
    \STATE Step env. to get recent reward $\boldsymbol{R}(t)$ and next state $\boldsymbol{S}(t+1)$
    \STATE Get EP reward ${R}_{ep}$ and MA reward ${R}_{ma}$
    \STATE Compute recent  reward-to-go  $\hat{\boldsymbol{R}}(t+1) \leftarrow \hat{\boldsymbol{R}}(t)-\boldsymbol{R}(t)$
    \STATE Update test trajectory $\tau_{t+1} \leftarrow (\tau_t, \hat{\boldsymbol{A}}(t), \hat{\boldsymbol{R}}(t+1), \boldsymbol{S}(t+1))$
    \ENDFOR
\STATE {\bfseries  Return:}   ${R}_{ep}$, ${R}_{ma}$ and the whole test trajectory $\tau_{{\small T}_{te}}$
\end{algorithmic}
\end{algorithm}

Furthermore, we construct stochastic prompts by randomly sampling a subsequence on the top-1 trajectory to increase model generalization. 
Formally, the training prompt $\tau_m^{(tr)\star}$ for the $m$-th user environment  consists of a trajectory segment of length $L_{pr}$ as follows:
 \begin{equation}
\begin{aligned}
&\tau_m^{(tr)\star}  
       = (\hat{\boldsymbol{R}}_i^{\star}, \hat{\boldsymbol{S}}_i^{\star}, \boldsymbol{A}_i^{\star},
       \cdots, 
       \hat{\boldsymbol{R}}_{i+L_{pr}}^{\star}, \hat{\boldsymbol{S}}_{i+L_{pr}}^{\star}, \boldsymbol{A}_{i+L_{pr}}^{\star}),\\
\end{aligned}
    \label{eq:prompt_representation_training}
\end{equation}
 where $i$ is randomly sampled between $[0,L-L_{pr})$. 

However, during online execution, FedPromptDT may encounter various user environments that are not encountered during its training phase. 
This means the MEC server cannot generate a trajectory prompt for the pre-trained FedPromptDT model by selecting a sample from its digit-MEC data.  
To address this issue, we  construct  the execution prompt $\tau_m^{(te)\star}$  with a length $L_{pr}$  as follows:
  \begin{equation}
\begin{aligned}
&\tau_m^{(te)\star}  
        = (\hat{\boldsymbol{R}}^{\star}, \hat{\boldsymbol{S}}(0), \boldsymbol{A}^{\star},
        \cdots,
        \hat{\boldsymbol{R}}^{\star}, \hat{\boldsymbol{S}}(0), \boldsymbol{A}^{\star}),
\end{aligned}
    \label{eq:prompt_representation_testing}
\end{equation}
where $\hat{\boldsymbol{R}}^{\star}$,    and $\boldsymbol{A}^{\star}$  are the target reward-to-go, the initial state augmented with user information and the user-preferred action on testing user environments, respectively.
The user-preferred action refers to fulfilling all user-preferred resolutions without regard to the MEC server's resource constraints.
For simplicity, we construct $\boldsymbol{A}^{\star}=\{\hat{r}_{e, k, a}, \hat{B}_{e, k},\hat{f}_{e, k}\}$, where $\hat{r}_{e, k, a}=1$, $\hat{B}_{e, k}=1$, and $\hat{f}_{e, k}=1$, $\forall a\in \{1,2,3\}$ and $\forall k \in \{1,2,\cdots,{K}_{e}\}$; otherwise, these elements are set as zeros  $\forall k \in \{{K}_{e},\cdots,{K}_{max}\}$.
The construction implies that each user prefers the MEC server to allocate as many resources as possible.

The pre-trained FedpromptDT model has learned how to generate the optimal path under resource constraints after offline training.
As a result,      $\boldsymbol{A}^{\star}$ in $\tau_m^{(te)\star}$ can guide the model to decide optimal user-preferred allocation without violating resource constraints. 
Meanwhile,  $\hat{\boldsymbol{S}}(0)$ in  $\tau_m^{(te)\star}$  motivates FedpromptDT to output an optimal user-preferred action as soon as possible after the initial environment state.
Note that we set the length of $\tau_m^{(te)\star}$ as $L_{pr}$ to eliminate the impact of inconsistent prompt lengths used in offline training and online execution on model prediction.

\subsubsection{FL-based offline training and online execution of FedPromptDT}\label{sec:OfflineFL_online_ex}

All MEC servers use FedAvg \cite{mcmahan2017communication} to train FedPromptDT during offline training according to Algorithm \ref{alg:FedPromptDT}.
Each MEC server locally minimizes the MSE loss $\mathcal{L}_{MSE}$ between the predicted actions $\hat{\boldsymbol{A}}_e$ and target actions $\boldsymbol{A}_e$   for both the prompt and training trajectories in its local dataset $\mathcal{D}_e$. 
The learning objective of our FedPromptDT framework is defined as follows:
\begin{equation}
 \begin{aligned}
   \min_{\mathbf{w} \in \mathbb{R}}  &  \mathcal{L}_{MSE}(\mathbf{w} )  =      \sum_{e=0}^{E-1} \frac{n_e}{n} 
 \mathcal{L}_{MSE}^{(e)}(\mathbf{w} )  \\
   & =    
   \sum_{e=0}^{E-1} \frac{n_e}{n} \mathbb{E}_{\mathcal{B}_{e,m}\subset \mathcal{D}_e}\left[
({\boldsymbol{A}}_{e} -\hat{\boldsymbol{A}}_{e})^2
   \right].
    \end{aligned}
   \label{eq:fl_promptDT}
   \vspace{-1mm}
\end{equation}
As shown in Algorithm \ref{alg:Localtraining},  the MEC server samples a user environment to get a batch that concatenates the prompt and training trajectories in each local iteration, i.e.,  $\mathcal{B}_{e,m}=(\tau_m^{(tr)\star},\tau_m)$.
 Along the training iterations $m\in [0,M)$, the server performs a batch gradient update on FedpromptDT iteratively on different user environments.
This motivates FedPromptDT to combine the $m$-th user environment information with recent training history for future action predictions.

In the execution evaluation phase, the pre-trained  FedPromptDT model interacts with the online environment to allocate resources for various user requests that are unseen in the training dataset, as shown in Algorithm \ref{alg:PromptDT_Eval}. 
At the beginning of the evaluation, a desired reward-to-go $\hat{\boldsymbol{R}}^{\star}$ and the initial environment state $\boldsymbol{S}(0)$ are provided for FedPromptDT as conditioning information.
Meanwhile,  the execution prompt representation, shown in (\ref{eq:prompt_representation_testing}), is designed to avoid sampling prompt trajectories from the training dataset.
Until the evaluation episode ends, FedPromptDT takes both the prompt and the latest context as input to autoregressively generate an action for the current user environment.



\section{Evaluation}
\label{sec:evaluation}
This section comprehensively evaluates the pre-trained FedPromptDT model, comparing it with baseline methods on user environments with varying user numbers and levels that are unseen during offline training.
After discussing the experimental settings, this section will present our performance evaluation on various user environments and MEC system settings, followed by ablation studies on  FedPromptDT.
\begin{table}[t]
\centering
\caption{Summary of simulation settings on MEC system and FL.}
\label{tab:simulation_setups}
\scalebox{0.82}
{
\begin{tabular}{@{}cc|cc@{}}
\toprule 
 Parameters & Value & Parameters & Value \\ \midrule
 \multicolumn{4}{c}{\textit{(MEC System Parameters)}}    \\ 
$F$ & 16 & $I, J, N$ & $4,4,16$ \\
$K$ & $[3,8]$ & $T_{t h}$ & $50 m s$ \\
$B_{max}$ & $10 M$ & $f_{max }$ & $15 \mathrm{GHz}$ \\
 $c_1$ & 800 cycles/bit & $c_2$ & 900 cycles/bit  \\
 $c_3$ & 1000 cycles/bit & $P_k$ & $1 W$ \\
 $b_{max}^{(\text{UHD})}$ & $12441600 \mathrm{bit}^1$ & $b_{max}^{(\text{FHD})}$ & $3110400 \mathrm{bit}^1$ \\
 $b_{max}^{(\text{HD})}$ & $1382400 \mathrm{bit}^1$ & 
 $b_{max}^{(\text{SD})}$ & $460800 \mathrm{bit}^1$ \\
  QoE$_{k,th}$ & 0.91 & hfQoE$_{th}$ & 0.8 \\ \midrule
   \multicolumn{4}{c}{(\textit{FL Parameters})}    \\   
   $E$ & 5 & $R$ & $100$ \\
 $\eta$  & $0.0001$ &    $B$ & $16$ (per user env.) \\
  $L_{tr}$  & $10$ &  $L_{pr}$ & $5$\\
    $M$  & $10$ &  Training env. no. & $20$ (per user no.) \\
     lr decay & $0.01$  (per round) &  Weight decay & $0.0001$ \\
   Optimizer & AdamW &  Warm-up step & $3$ (per round) \\
\bottomrule
\end{tabular}
}
 \end{table}

\begin{table}[t]
\centering
\caption{Summary of FedPromptDT Architectural Setting.}
\label{tab:PromptDT_setups}
\scalebox{0.88}
{
\begin{tabular}{@{}cc|cc@{}}
\toprule 
 Parameters & Value & Parameters & Value \\ \midrule   
Attention layer  & $6$ &  Head number & $1$ \\
 Dropout & $0.1$ &  Embedding dimension & $128$  \\
Input dimension &  $98$ & Output dimension & $40$  \\
Transformer activation & ReLU & Head activation & Sigmoid   \\
\bottomrule
\end{tabular}
}
 \end{table}
\begin{table*}[t]
\centering
\caption{Performance evaluation on ten evaluation episodes in two scenarios.
We report the results in the mean (standard deviation) format to describe performance on the ten episodes.
Bold and underlined text indicate the best mean and standard deviation results among all the methods, respectively. }
\label{tab:main_results}
\scalebox{0.95}{
\begin{tabular}{@{}ccccccc@{}}
\toprule
\multirow{2}{*}{Methods} & \multicolumn{3}{c}{Scenario 1 ($\forall K \in \{3,4\}$, Env. No. = 20)}                  & \multicolumn{3}{c}{Scenario 2 ($\forall K\in \{3,4,\cdots,8\}$, Env. No. = 60)}                   \\ \cmidrule(l){2-7} 
                         & MA Rewards          & EP Rewards             & Min QoE             & MA Rewards          & EP Rewards              & Min QoE             \\ \midrule
Local DDPG ($K=4$)         & 9.24(0.22)          & 931.73(17.71)          & 1.09(0.05)          & 5.84(2.55)          & 588.10(259.11)          & 0.76(0.44)          \\
Local DDPG ($K=8$)         & 8.28(0.42)          & 834.75(25.05)          & 1.04(0.10)          & 6.59(1.67)          & 664.78(161.05)          & 0.67(0.66)          \\
Local DT                 & 9.57(0.23)          & 946.37(14.74)          & 1.10(0.05) & 7.45(2.40)          & 704.99(253.10)          & 0.71(0.45)          \\
Local PromptDT           & 9.50(0.21)          & 934.48(\underline{11.18})          & 1.01(0.36)          & 8.03(1.93)          & 816.97(136.62)          & 0.76(0.48)          \\ \midrule
FedDDPG ($K=4$)            & 9.41(0.28)          & 940.62(37.91)          & 1.01(0.28)          & 6.41(2.43)        &649.58(228.80)           & 0.79(0.32)          \\
FedDDPG ($K=8$)            & 9.17(0.23)          & 918.83(21.74)          & 1.09(\underline{0.03})          & 7.69(\underline{1.38}) & 772.88(139.71)          & 0.82(0.36)          \\
SMoE FedDDPGs           & 9.37(0.39)          & 947.44(15.72)          & \textbf{1.14}(0.07)          & 8.02(1.52)          & 812.06(142.87)          & \textbf{0.89}(\underline{0.26})          \\ \midrule
FedDT                    & 9.35(0.43)          & 944.62(19.12)          & 1.09(0.04)          & 7.46(2.41)         & 693.60(254.64)         & 0.61(0.37)         \\
FedPromptDT              & \textbf{9.59}(\underline{0.14}) & \textbf{950.33}(17.32) & 1.09(0.05)          & \textbf{8.20}(1.57) & \textbf{829.61}(\underline{128.74}) & 0.84(0.32)\\ \bottomrule
\end{tabular}
}
\end{table*}

\subsection{Experimental Settings}

\subsubsection{Simulation setting}
We consider a MEC system with $5$ MEC servers to conduct FL to obtain the pre-trained FedPromptDT model.
These MEC servers have the same capacity for computation and communication. 
Each MEC server is located at the origin of the coordinates
 (i.e., $[0m, 0m]$) with its VR users to communicate.
The $x$-coordinate and $y$-coordinate of these users are randomly changed within $[10m,20m]$  and $[0m,5m]$, respectively. 
The distance-dependent path-loss exponent $\alpha = 4$. the channel noise power is ${\sigma}^{2}   = -174 dBm$, and the compression ratio before transmission $\omega=300$.
Meanwhile, MEC provides three user  levels   with attention-based resolution thresholds, including:
\begin{itemize}
    \item Premium user: ${b}_{k,1,th}^{(\text{pr})}=b_{max}^{(\text{HD})}/2$,  ${b}_{k,2,th}^{(\text{pr})}=b_{max}^{(\text{FHD})}/2$, ${b}_{k,3,th}^{(\text{pr})}=b_{max}^{(\text{UHD})}/4$;
    \item Advanced user: ${b}_{k,1,th}^{(\text{ad})}=b_{max}^{(\text{SD})}/1.5$, ${b}_{k,2,th}^{(\text{ad})}=b_{max}^{(\text{HD})}/2$, ${b}_{k,3,th}^{(\text{ad})}=b_{max}^{(\text{FHD})}/2$; 
    \item Standard user:
    ${b}_{k,1,th}^{(\text{st})}=b_{max}^{(\text{SD})}/2$, ${b}_{k,2,th}^{(\text{st})}=b_{max}^{(\text{SD})}/1.5$,  ${b}_{k,3,th}^{(\text{st})}=b_{max}^{(\text{HD})}/2$.
\end{itemize}
Table \ref{tab:simulation_setups} summarizes the parameter values used for system simulation and FL, where the subscript $e$ of the MEC index is omitted for simplicity.
Our simulation code is implemented using PyTorch and the Huggingface Transformers library.
  All experiments are performed based on two nodes of a High-Performance Computing platform with 8 NVIDIA A30 Tensor Core GPUs with 24GB per node.

\subsubsection{FedPromptDT architecture setting}
We follow  the model architecture from \cite{Chen2021decision,xu2022PromptDT} and design the prompt as (\ref{eq:prompt_representation_training}) and (\ref{eq:prompt_representation_testing}), as shown in Figure \ref{fig:systemmodel}.   
Specifically, the model learns three linear layers to project raw inputs to the desired embedding dimension, followed by layer normalization. An additional embedding is learned and added to each token for each timestep.
Table \ref{tab:PromptDT_setups} summarizes the FedPromptDT architecture settings.

 \subsubsection{Baseline methods}
We show the performance advantages of FedPromptDT from three aspects: one is to compare the impact of local training and  FL on PromptDT; the second is to compare with the value policy-based RL; the third is to demonstrate the effect of the prompt.
Our baseline methods are summarized as follows:
\begin{itemize}
    \item Local DDPG: Each MEC server locally conducts DDPG to learn a behavior policy for its local environments based on user number $K$, e.g., referred to Local DDPG (K=4) when $K=4$.
    \item Local DT (or PromptDT): Each MEC server locally trains a DT model without (or with)  prompts with its local data.
    \item Federated DDPG (FedDDPG): All MEC servers use FL to learn a behavior policy for the environments of a user number $K$, where the critic network and its target network are shared and averaged during FL training, e.g., referred to FedDDPG (K=4) when $K=4$.
    \item Selective model ensemble of FedDDPGs (SMoE FedDDPGs): We perform a selected output ensemble of FedDDPG actor networks, e.g., when given a user environment $K=5$, SMoE FedDDPG ensembles the output of FedDDPG actor networks of $K \in \{5,6,\cdots,8\}$. 
    \item Federated DT (FedDT):  All MEC servers use FL to train a DT model for all user environments without the prompt.
    \item FedDT with fine-tuning: Each MEC server takes its evaluation memories to fine-tune the pre-trained FedDT model without the prompt.
\end{itemize}

 \subsubsection{Dataset preparation and pre-trained models}
The attention level data is processed based on the collected eye gaze data\footnote{https://github.com/xuyanyu-shh/VR-EyeTracking} from \cite{Xu2018video360}. 
Specifically, the study \cite{Xu2018video360} utilized 208 dynamic 360° videos from YouTube, featuring indoor and outdoor scenes, music shows, sports games, documentaries, and short movies.
45 participants wore head-mounted devices integrated with an eye tracker to play these video clips and capture the viewers’ gaze, resulting in users' eye gaze records.
Note that we excluded 5 participant records for being too short and used the remaining 40 participant records.

In our experiments, the eye-gaze records of each user are divided into training and testing records.
We randomly combined the eye gaze coordinate data from each user's training (testing) records to create attention-based training (testing)  user data, where tile numbers are based on $F=16$ frames of gaze center coordinates.
We have a total of 5 MEC servers, where each  MEC server's training and testing dataset accommodates up to 8 users' data due to a total of 40 participants.
We first train our baseline DDPG to learn a behavior policy for the user environments with a user number $K$  ($\forall K \in \{3,4,\cdots,8\}$).
We then collect and combine its training memory stored in digital MECs as its offline dataset for pre-training FedPromptDT, consisting of approximately 100 to 200 trajectories per environment. 
Finally, we take all the data in digital MECs to conduct the pre-training on FedPromptDT according to Algorithm \ref{alg:FedPromptDT}.

\subsubsection{Evaluation metric}
The total number of user environments equals the product of the user number and user level.
We consider $10$ user levels unseen in the training dataset on the user environment of each user number.
We evaluate the experiments using the following metrics: accumulated episode (EP) reward, moving averaging (MA) reward, and minimal QoE along the whole episode. 
The  EP and MA rewards measure the performance of allocation decisions based on the achieved QoE and hfQoE throughout the whole episode and at one step, respectively.
The minimal QoE measures whether allocation decisions meet the QoE threshold ($\text{QoE}_{k,th} = 0.91$).
Meanwhile,  we choose the initial reward-to-go as 900 for FedPromptDT in the test evaluation.

\subsection{Performance Evaluation}

\begin{figure}[t]
	\centering  
	\subfigcapskip=-2pt 
    \hspace{-5mm}
    	\subfigure[Different env. user numbers]{
		\includegraphics[width=0.245\textwidth]{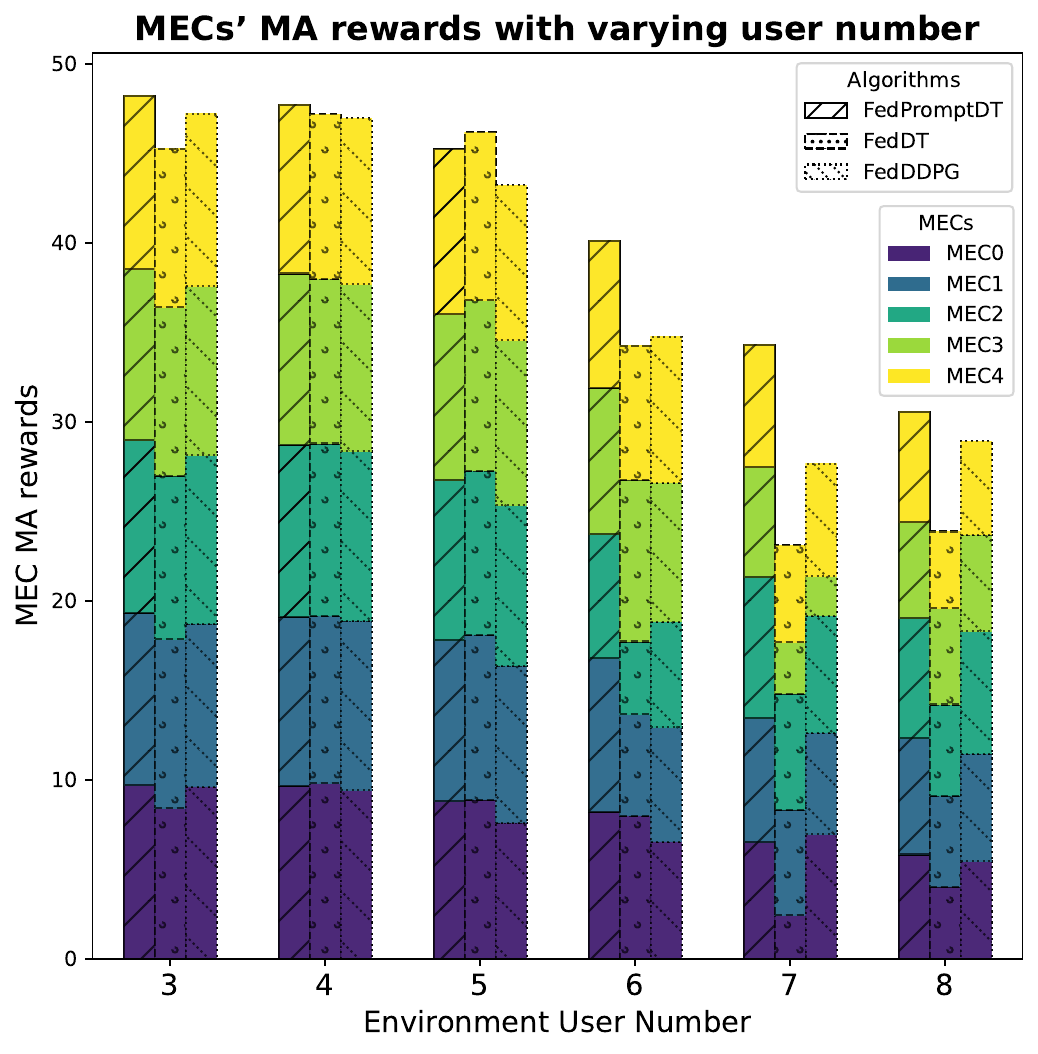}
    \label{fig:varying_user_number}}
    \hspace{-5mm}
	\subfigure[Different user level ($K=8$)]{
		\includegraphics[width=0.244\textwidth]{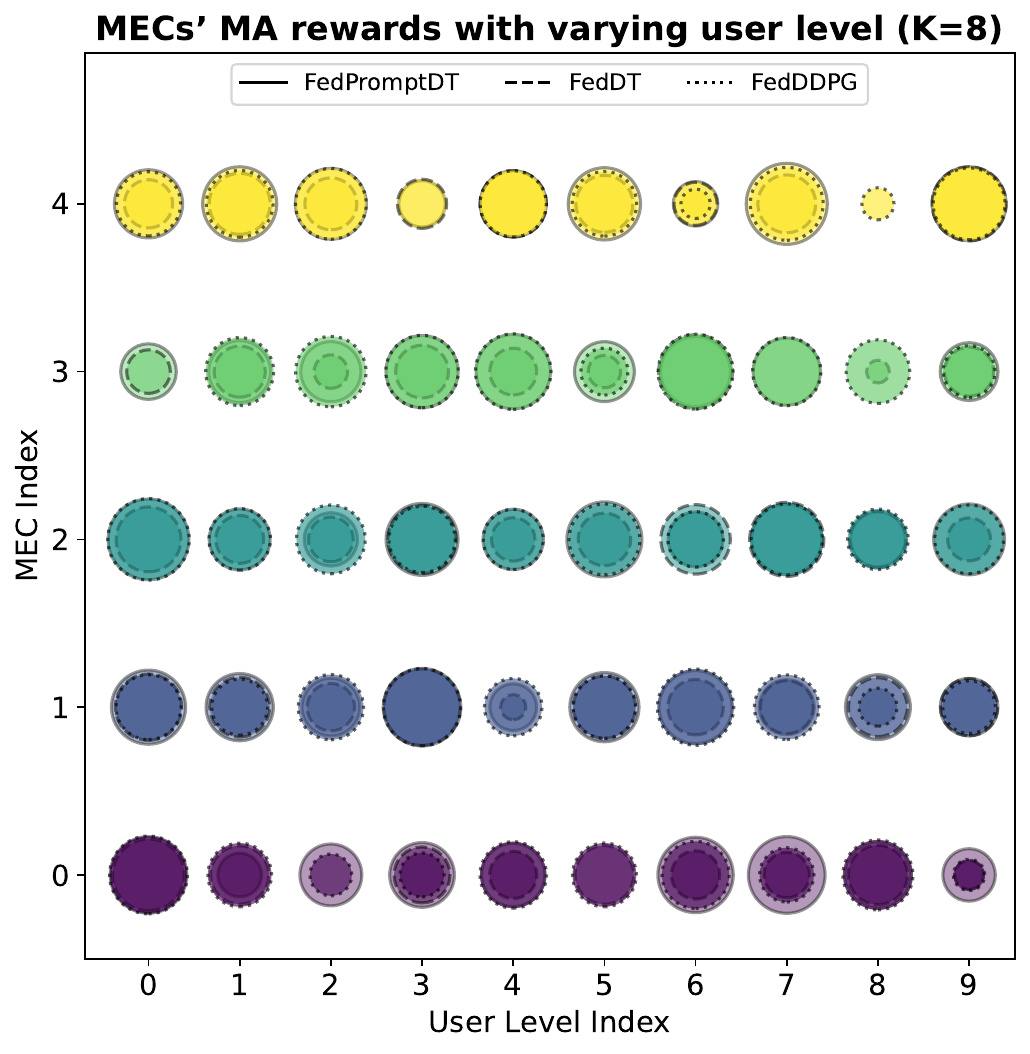}
    \label{fig:varying_user_level}}
  \caption{(a): MEC MA rewards on different user numbers; (b): MEC MA rewards on different user levels.}
  \label{fig:comprehensizve_ma_rewards}
\end{figure}

\subsubsection{Main results}
The results displayed in Table \ref{tab:main_results} demonstrate the superior performance of FedPromptDT compared to our baselines. 
 Table \ref{tab:main_results} includes two test scenarios:  varying user number $\forall K \in \{3,4\}$ and $\forall K\in \{3,4,\cdots,8\}$.
User environments involve $10$ user levels unseen during training for each user number.

Conducting FL enhances the PromptDT performance on MEC servers compared to local training.
In both scenarios, FedPromptDT achieves higher mean rewards and  QoE than local PromptDT while reducing the performance instability, i.e., smaller standard deviation of MA rewards and minimal QoE on various user environments.
However, FL does not improve the DT performance.
This indicates that MECs' local policies are inconsistent with the expected policy.
In other words, FedDT learns only the policy specific to its local data during local training due to the lack of inductive bias to distinguish different user environments.
Moreover,  FL improves the DDPG performance since DDPG is solely trained on user environments with a given user number.
There is no need for DDPG to differentiate between user environments based on the number of users.

Furthermore, FedPromptDT demonstrates strong generalization capabilities across a wide range of user environments, aided by the use of the prompt.
Compared to FedDT,  FedPromptDT shows higher rewards and stability in various environments; e.g.,   in Scenario 2, FedPromptDT achieved a mean episode reward of 829.61, surpassing FedDT by 136.01 rewards, and with half the standard deviation.
Compared to FedDDPG, FedPromptDT exhibits superior performance across environments with varying user numbers, as well as consistent performance benefits under the case of only evaluating user environments related to  FedDDPG's trained user numbers as per Figure \ref{fig:varying_user_number}.
Additionally,  FedPromptDT exhibits comparable performance to SMoE FedDDPGs, which ensembles the output of actor networks on various FedDDPGs, indicating the low inference cost of FedPromptDT.
 \begin{figure}[t]
	\centering  
	\subfigcapskip=-2pt 
    \hspace{-5mm}
    	\subfigure[Different thresholds in ($\textbf{P}_0$)]{
		\includegraphics[width=0.245\textwidth]{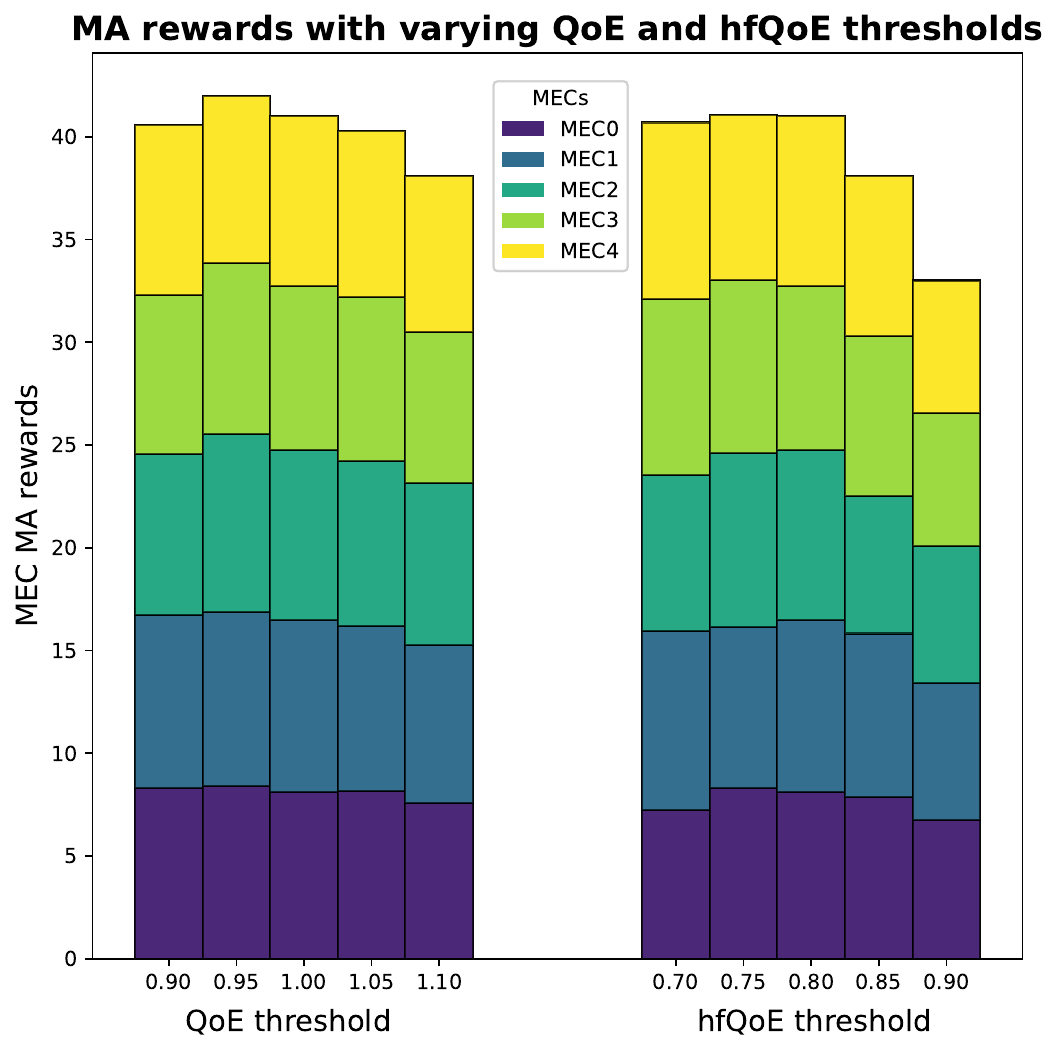}
    \label{fig:varying_thresholds}}
    \hspace{-5mm}
	\subfigure[Different MEC settings]{
		\includegraphics[width=0.244\textwidth]{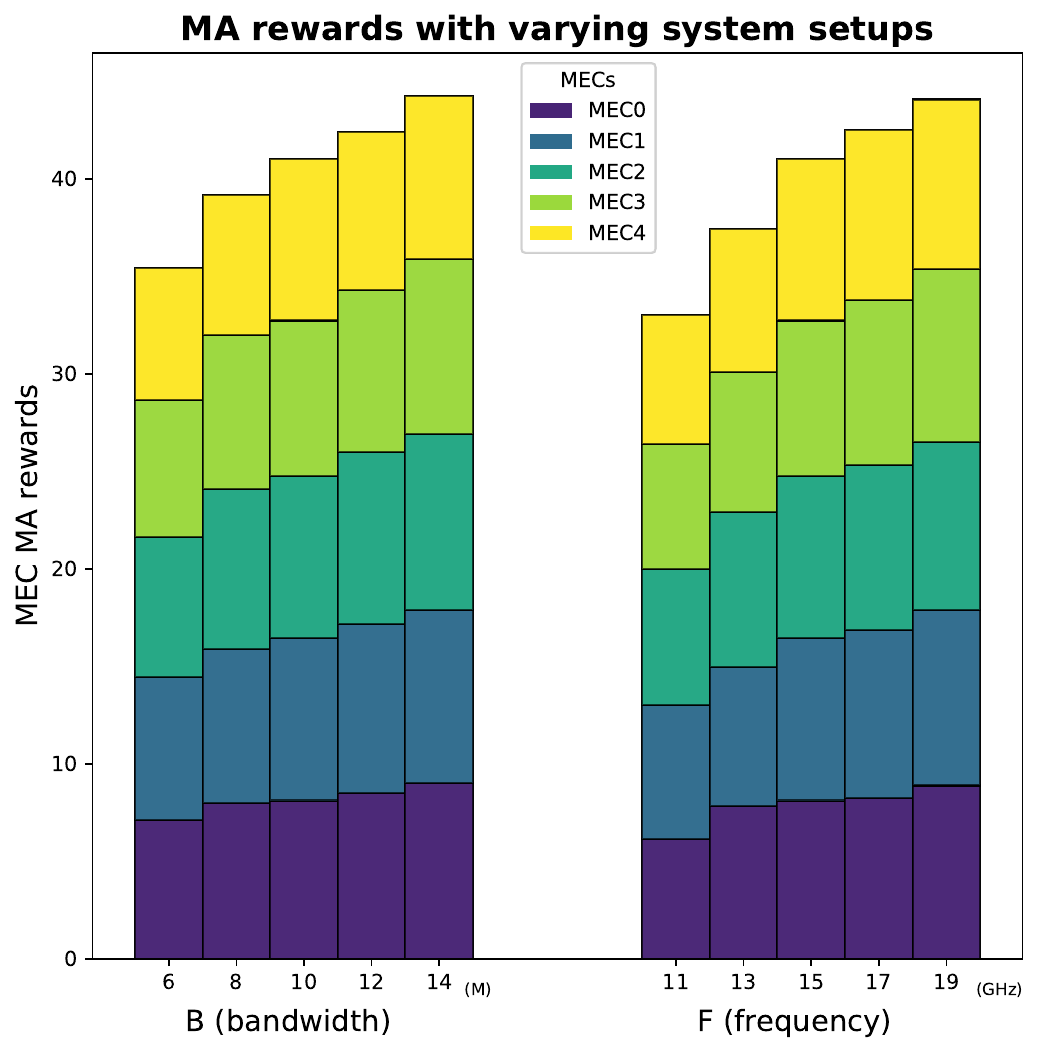}
    \label{fig:varying_sys_params}}
  \caption{(a): MEC MA rewards on different QoE and hfQoE thresholds in ($\textbf{P}_0$); (b): MEC MA rewards on different MEC bandwidth and frequency capability. }
  \label{fig:different_threshold}
\end{figure}

\begin{table}[t]
\centering
\caption{Comparison between prompt-based and FT-based effectiveness on DT and PromptDT.
We conduct testing on ten evaluation episodes of the user environments ($\forall K \in \{3,4,\cdots,8\}$) and report the results in the format of the mean (standard deviation).
\textit{G} and \textit{L} refer to the global and local models from the final round of FL, respectively. }
\scalebox{0.88}
{
\begin{tabular}{@{}cccc@{}}  
\toprule
\multirow{2}{*}{Methods} & \multicolumn{3}{c}{Senario 2 (K=8, Env. No.=60)} \\ \cmidrule(l){2-4} 
                         & MA Rewards    & EP Rewards        & Min QoE      \\ \midrule
FedDT(No Prompt)         & 7.46(2.41)    & 693.60(254.64)    & 0.61(0.37)   \\
FedPromptDT (\textit{G})      & \textbf{8.29}(\underline{1.40})    & 829.75(\underline{125.22})    & 0.81(0.37)   \\
FedPromptDT (\textit{L})            & 8.20(1.57)    & 829.61(128.74)    & \textbf{0.84}(\underline{0.32})   \\ \midrule
FedDT w. FT              & 6.99(2.94)    & 676.46(295.26)    & 0.54(0.41)   \\
FedPromptDT w. FT        & \textbf{8.29}(1.72)    & \textbf{834.30}(130.21)    & 0.82(0.39)   \\ \bottomrule
\end{tabular}
}%
\label{tab:ablation_study}
\end{table}

\subsubsection{Performance on various user numbers and user levels}

 To further show the effectiveness of FedPromptDT,  we compare it with FedDT and FedDDPG under varying user numbers and user levels, as shown in Figure \ref{fig:comprehensizve_ma_rewards}, where
FedDDPG's results are obtained from training on the given user number.

 Figure \ref{fig:varying_user_number} illustrates  MECs' MA rewards achieved by each method when given a user number. 
 FedPromptDT outperforms FedDDPG and FedDT (except for $K=5$) across all user numbers, showcasing its superior adaptability to various user environments.
 Moreover, compared to FedDT, FedPromptDT exhibits a much wider performance gap on $K\geq6$ than on $K\leq5$, highlighting its ability to handle more complex scenarios with a larger user base.

Figure \ref{fig:varying_user_level} displays MECs' MA rewards on ten user environments depending on different user levels and a fixed user number of $K=8$, where the larger circle indicates the higher MA reward.
FedPromptDT typically achieves higher or comparable MA rewards across all user levels at each MEC, emphasizing the stability in handling various user levels.

\subsubsection{Performance on various QoE and hfQoE thresholds}

To explore the redundancy of output action to various thresholds of $\textbf{P}_0$, we use a pre-trained FedPromptDT to work with different QoE and hfQoE thresholds based on Scenario 2 of Table \ref{tab:main_results}, as shown in Figure \ref{fig:varying_thresholds}.
 The figure shows the results of FedPromptDT in terms of MA rewards at each MEC server over a range of QoE thresholds (i.e., $\text{QoE}_{e,k,th}$ from 0.90 to 1.10 with a step of 0.05) and hfQoE thresholds (i.e., $\text{hfQoE}_{e,th}$ from 0.70 to 0.90 with a step of 0.05), where $\text{QoE}_{e,k,th}=1.00$ and $\text{hfQoE}_{e,th}=0.8$ is the pre-trained setting of FedPromptDT.
 The pre-trained FedPromptDT maintains similar MA rewards across these tested thresholds, except for the most stringent threshold $\text{hfQoE}_{e,th}=0.9$.
This suggests that the output of the pre-trained FedPromptDT has sufficient redundancy to accommodate various threshold settings. 
 
\subsubsection{Performance on various system settings}
 We also investigate how the output actions of the pre-trained FedPromptDT are affected by varying MEC system settings, as shown in Figure \ref{fig:varying_sys_params}.
  The figure illustrates FedPromptDT's performance in terms of MA rewards when applied to different MEC frequencies and bandwidths. 
The range of bandwidth $B_{e,max}$  is from 6Mbps to 14Mbps with a step of 2Mbps, and the range of frequency  $f_{e,max}$ is from 11GHz to 19GHz  with a step of 2GHz, where $B_{e,max}=10$Mbps and $f_{e,max}=15$GHz is the pre-trained setting. 
The pre-trained FedPromptDT model achieves higher MA rewards as the bandwidth and frequency increase.
 The observed results are consistent with our expectations, suggesting that the model has learned a policy that is not limited to specific system settings but can accommodate different ones.
 
\begin{figure}[t]
	\centering  
	\subfigcapskip=-2pt 
    \hspace{-5mm}
    	\subfigure[Different $L_{pr}$ and data sizes]{
		\includegraphics[width=0.245\textwidth]{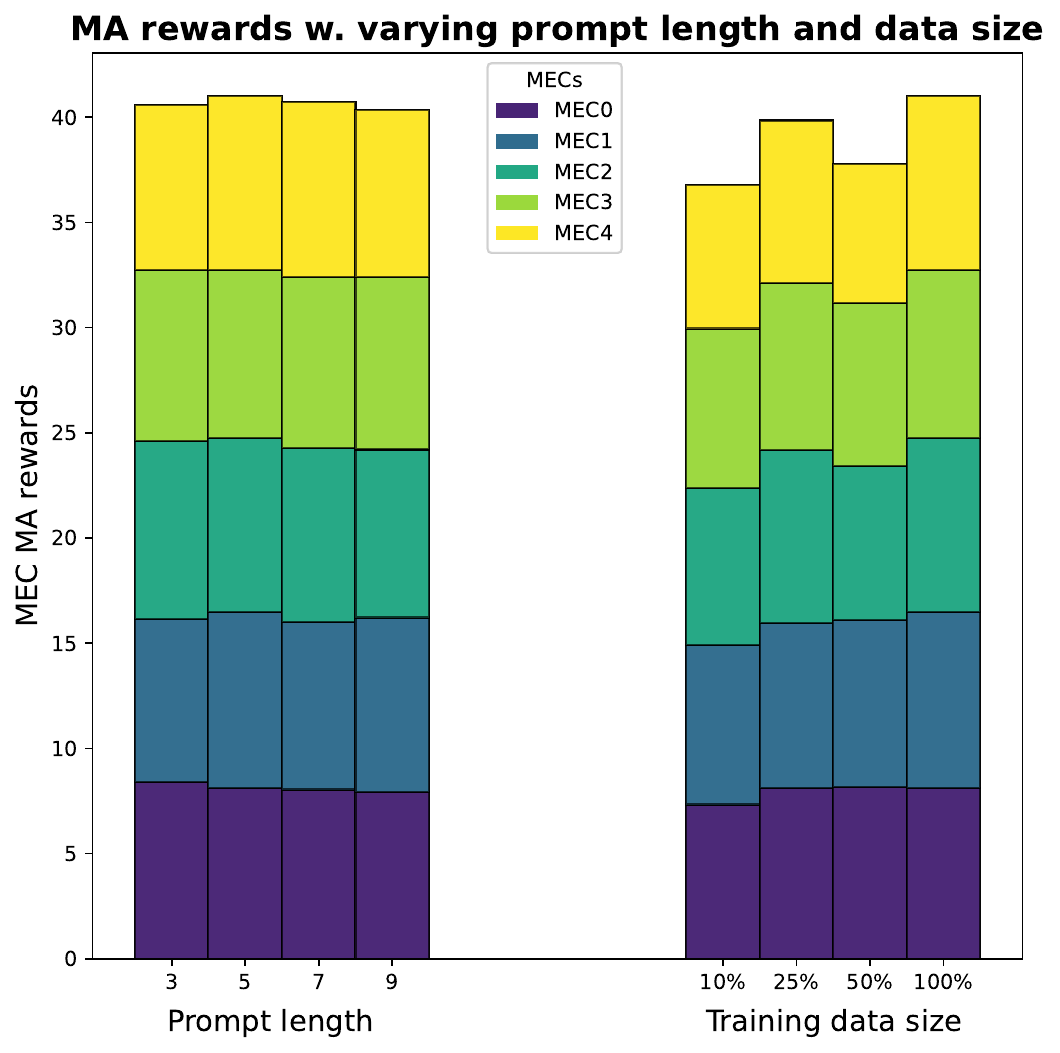}
    \label{fig:varying_prompt_len_datasize}}
    \hspace{-5mm}
	\subfigure[Different FL training settings]{
		\includegraphics[width=0.24\textwidth]{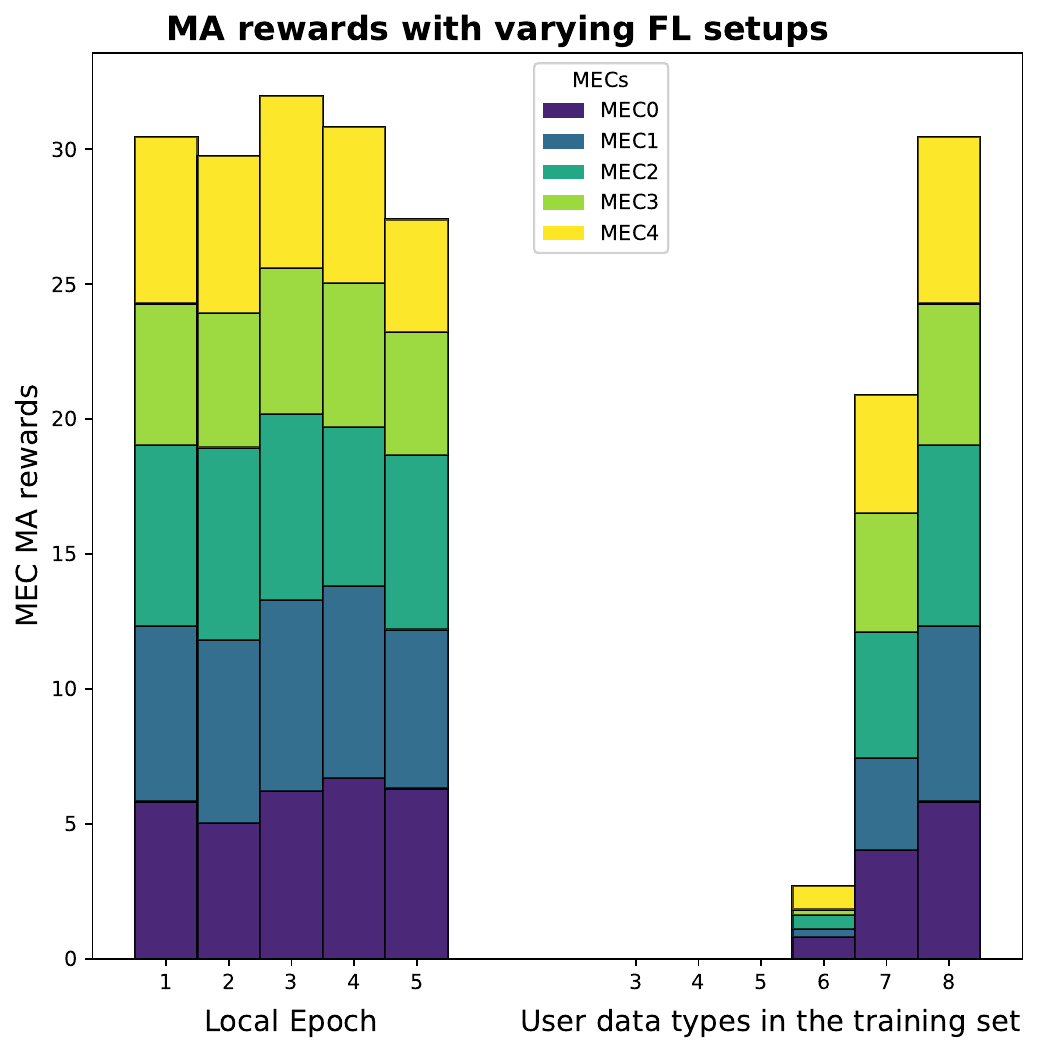}
    \label{fig:varying_FL_setups}}
  \caption{(a): MEC MA rewards on different prompt length $L_{pr}$ and various training data sizes in FL; (b): MEC MA rewards when using different local epochs and user environments to perform FL training. We categorize the user environment into user data types based on user number.}
  \label{fig:prompt_FL_setups}
\end{figure}

\subsection{Ablation Studies}
\subsubsection{Comparison between Prompt and  Fine-tuning (FT)}
FT trains pre-trained models with a small dataset to improve performance for specific use cases.
 In this work, we use the trajectory data of the episode that yields maximum rewards in previous episodes as an FT dataset during evaluation.
We fine-tuned FedDT and FedPromptDT on the user environments  ($\forall K \in \{3,4,\cdots,8\}$) with ten FT iterations and report their performance in  Table \ref{tab:ablation_study}.

The table shows the performance of FedDT with FT is not better than FedDT without FT.
This suggests that the effect of FT on DT  depends on the generalization capability of the pre-trained DT model.
In contrast, FedPromptDT with FT yields higher EP rewards than FedPromptDT without FT, while its MA rewards and QoE remain comparable.
This demonstrates that the prompt design allows FedPromptDT to generalize well across various user environments even without FT.
Moreover, the powerful generalization of FedPromptDT can bring gains when performing FT on it.
We also evaluate the impact of prompt design on FedPromptDT using different prompt lengths ($L_{pr}$ from 3 to 10 with a step of 2), as illustrated in Figure \ref{fig:varying_prompt_len_datasize}.
The results demonstrate that the length of the prompt does not affect FedPromptDT's performance.

\subsubsection{The effect of FL settings on pre-training FedPromptDT}
We vary the FL training settings, including different training dataset sizes, local epochs, and user environment types used during training, to generate various pre-trained FedPromptDT models. 
We then evaluate the performance of these models in terms of MEC MA rewards based on the user environments $K \in \{3,4,\cdots,8\}$ in Figure \ref{fig:varying_prompt_len_datasize} and $K =8$ in Figure \ref{fig:varying_FL_setups}.

Firstly, we train FedPromptDT on various dataset sizes, where the full size comprises 20 user environments per user, with around 100 to 200 trajectories per environment.
Figure \ref{fig:varying_prompt_len_datasize} indicates that FedPromptDT does not require a large training dataset size.
Using only $10\%$ of the data can result in a decent pre-trained FedPromptDT, while $25\%$  data can achieve a comparable model to the full dataset. 
Secondly, given 100 FL communication rounds, we vary the local training epochs ranging from 1 to 5 for pre-training, with ten local iterations per epoch.
Figure \ref{fig:varying_FL_setups} demonstrates that using three local epochs for FedPromptDT yields the best results, achieving the optimal trade-off between utility and communication.
Thirdly, we categorize the user environment based on user number, referred to as user data type in Figure \ref{fig:varying_FL_setups}, and conduct FL training to obtain pre-trained FedPromptDT models for different user data types.
The figure shows that the models trained with the user data type $K<7$ struggle to generalize to the user environments $K=8$. 
This, coupled with the findings of Figure \ref{fig:varying_prompt_len_datasize}, suggests that the number of user environments used in FedPromptDT's pre-training can be small, while the environment diversity should be enriched.

\begin{figure}[t]
	\centering  
	\subfigcapskip=-2pt 
    \hspace{-5mm}
    	\subfigure[EP rewards]{
		\includegraphics[width=0.245\textwidth]{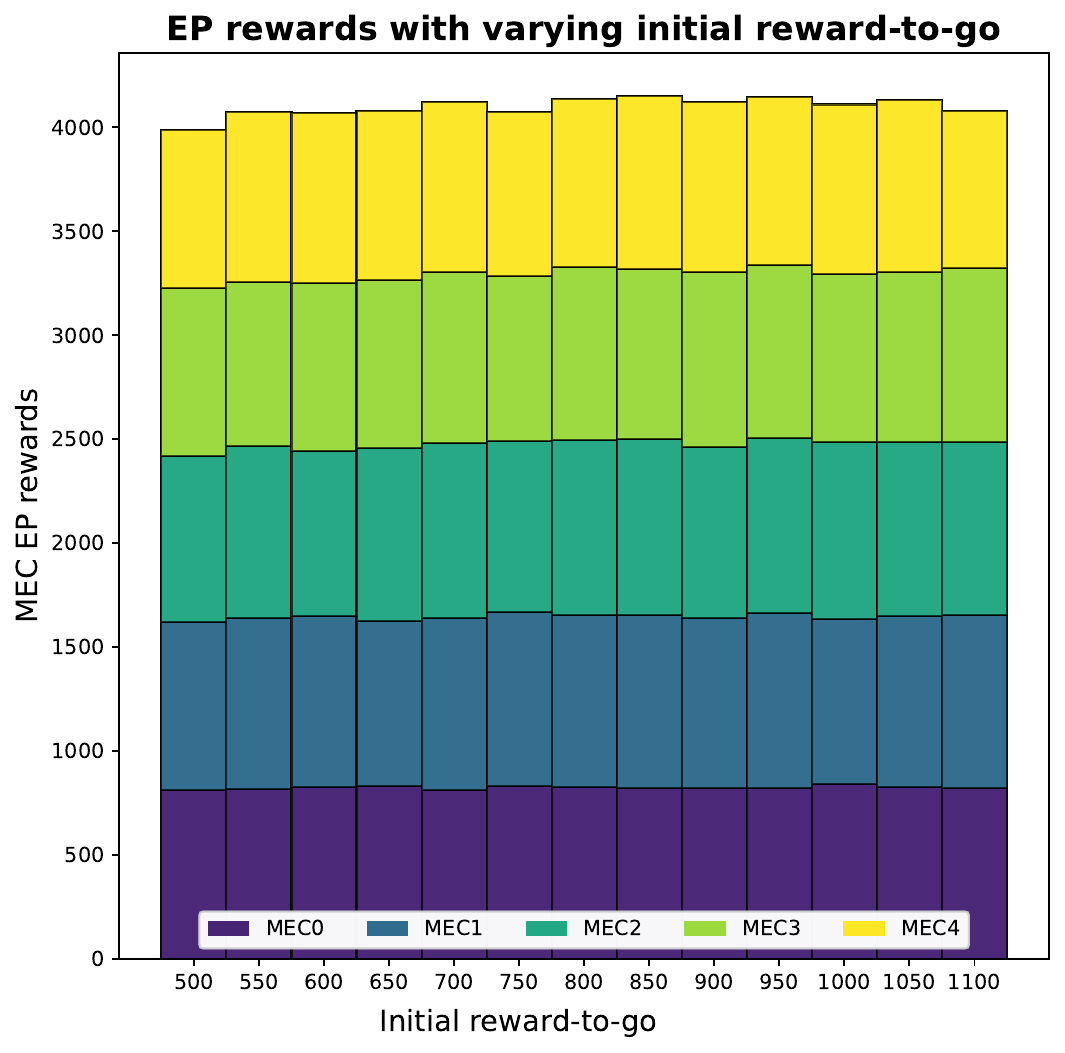}
    \label{fig:EP_initial_rtg}}
    \hspace{-5mm}
	\subfigure[MA rewards]{
		\includegraphics[width=0.239\textwidth]{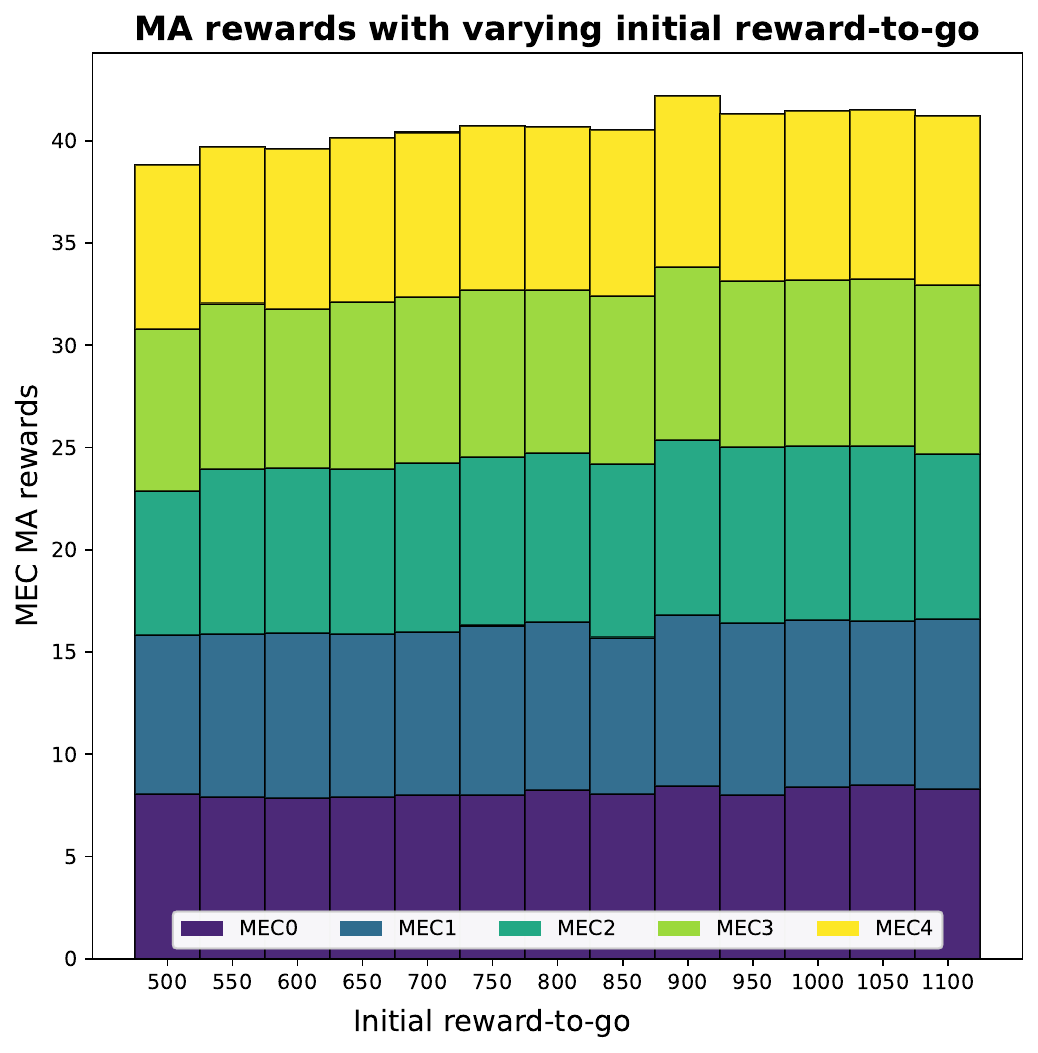}
    \label{fig:MA_initial_rtg}}
  \caption{(a): MEC EP rewards on different initial target reward-to-go; (b): MEC MA rewards on different initial target reward-to-go.}
  \label{fig:initial_rtg}
\end{figure}
\subsubsection{The effect of initial reward-to-go on FedPromptDT}
 As shown in Figure \ref{fig:initial_rtg}, our experiments investigate the effect of initial reward-to-go on the pre-trained FedPromptDT, varying from 500 to 1100 with a step of 50. 
 FedPromptDT consistently achieved similar EP and MA rewards across the different settings, i.e., achieving about 800 EP rewards and 8 MA rewards per MEC. 
  The results indicate that the performance of FedPromptDT is not significantly affected by the initial reward-to-go settings. 
 This provides FedPromptDT with more flexibility and adaptability when it comes to implementing it in different user environments.

\section{Conclusion}
\label{sec:conclusion}

This paper presented a FedPromptDT framework to address the challenge of resource allocation when the MEC system provides customized VR services for heterogeneous users. 
To evaluate the immersive experience for VR users, we first introduced a customized QoE metric that combines the MEC system latency, user attention levels, and user-preferred resolutions.
By optimizing the allocation of CPU frequency, bandwidth resources, and customized resolution, we formulated an attention-based QoE maximization problem under constraints on QoE and hfQoE constraints to enhance the QoE.
Next,  we transformed the problem into an RL problem to learn a generalized policy for various user environments across all MEC servers.
Our proposed FedPromptDT framework utilizes prompt-based sequence modeling to learn the policy.
It leverages FL to pre-train a FedPromptDT model and incorporates prompt design to inform the model with environmental information and user-preferred allocation decisions.
 With the benefits of the prompt design, FedPromptDT can easily adapt to different user environments and ensure effective allocation decisions for customized user requirements without re-training. 
We conducted extensive experiments on performance evaluation and ablation study on FedPromptDT. 
FedPromptDT achieved consistently superior performance in various user environments compared to our baselines. 

In summary, this paper introduced FedPromptDT as a generalized decision model for various environments on MEC's resource allocation.
This demonstrated its potential as a powerful and scalable solution for resource allocation.
In future work, it would be valuable to explore the potential of this framework in addressing resource allocation problems beyond our customized VR services.
Meanwhile, it would be interesting to incorporate a semantic prompt generator into this framework instead of our trajectory prompts. 
This generator can take in environment descriptions with texts to create prompts for the decision model, improving the model's interpretability and scalability for various environments.

\bibliographystyle{IEEEtran}
\bibliography{IEEEabrv,mybib}

\begin{thebibliography}{10}
\providecommand{\url}[1]{#1}
\csname url@samestyle\endcsname
\providecommand{\newblock}{\relax}
\providecommand{\bibinfo}[2]{#2}
\providecommand{\BIBentrySTDinterwordspacing}{\spaceskip=0pt\relax}
\providecommand{\BIBentryALTinterwordstretchfactor}{4}
\providecommand{\BIBentryALTinterwordspacing}{\spaceskip=\fontdimen2\font plus
\BIBentryALTinterwordstretchfactor\fontdimen3\font minus \fontdimen4\font\relax}
\providecommand{\BIBforeignlanguage}[2]{{%
\expandafter\ifx\csname l@#1\endcsname\relax
\typeout{** WARNING: IEEEtran.bst: No hyphenation pattern has been}%
\typeout{** loaded for the language `#1'. Using the pattern for}%
\typeout{** the default language instead.}%
\else
\language=\csname l@#1\endcsname
\fi
#2}}
\providecommand{\BIBdecl}{\relax}
\BIBdecl

\bibitem{WangSZXLLS23Metaverse}
Y.~Wang, Z.~Su, N.~Zhang, R.~Xing, D.~Liu, T.~H. Luan, and X.~Shen, ``A survey on metaverse: Fundamentals, security, and privacy,'' \emph{{IEEE} Commun. Surv. Tutorials}, vol.~25, no.~1, pp. 319--352, Oct. 2023.

\bibitem{ChangKY20VRSickness}
E.~Chang, H.~T. Kim, and B.~Yoo, ``Virtual reality sickness: {A} review of causes and measurements,'' \emph{Int. J. Hum. Comput. Interact.}, vol.~36, no.~17, pp. 1658--1682, Oct. 2020.

\bibitem{MaoYZHL17}
Y.~Mao, C.~You, J.~Zhang, K.~Huang, and K.~B. Letaief, ``A survey on mobile edge computing: The communication perspective,'' \emph{{IEEE} Commun. Surv. Tutorials}, vol.~19, no.~4, pp. 2322--2358, Aug. 2017.

\bibitem{XuNLKXNYSM23}
M.~Xu, W.~C. Ng, W.~Y.~B. Lim, J.~Kang, Z.~Xiong, D.~Niyato, Q.~Yang, X.~Shen, and C.~Miao, ``A full dive into realizing the edge-enabled metaverse: Visions, enabling technologies, and challenges,'' \emph{{IEEE} Commun. Surv. Tutorials}, vol.~25, no.~1, pp. 656--700, Oct. 2023.

\bibitem{yu2022bi}
J.~Yu, A.~Alhilal, P.~Hui, and D.~H.~K. Tsang, ``Bi-directional digital twin and edge computing in the metaverse,'' [Online]. Available: \url{https://arxiv.org/pdf/2211.08700.pdf}.

\bibitem{yu2023attention}
J.~Yu, A.~Alhilal, T.~Zhou, H.~Pan, and D.~H.~K. Tsang, ``Attention-based qoe-aware digital twin empowered edge computing for immersive virtual reality,'' [Online]. Available: https://arxiv.org/pdf/2305.08569.pdf.

\bibitem{8717893}
J.~v. der Hooft, M.~Torres~Vega, S.~Petrangeli, T.~Wauters, and F.~D. Turck, ``Optimizing adaptive tile-based virtual reality video streaming,'' in \emph{IFIP/IEEE Symp. Integr. Netw. Serv. Manage. (IM)}, Washington, DC, USA, Apr. 2019, pp. 381--387.

\bibitem{8626197}
C.~Ozcinar, J.~Cabrera, and A.~Smolic, ``Visual attention-aware omnidirectional video streaming using optimal tiles for virtual reality,'' \emph{IEEE J. Emerg. Sel. Topics Circuits Syst.}, vol.~9, no.~1, pp. 217--230, Jan. 2019.

\bibitem{li2023RLTransformers}
W.~Li, H.~Luo, Z.~Lin, C.~Zhang, Z.~Lu, and D.~Ye, ``A survey on transformers in reinforcement learning,'' \emph{Trans. Mach. Learn. Res.}, Sep. 2023.

\bibitem{radford2018improving_GPT1}
A.~Radford, K.~Narasimhan, T.~Salimans, I.~Sutskever \emph{et~al.}, ``Improving language understanding by generative pre-training,'' [Online]. Available: \url{https://www.mikecaptain.com/resources/pdf/GPT-1.pdf}.

\bibitem{Brown2020GPT3}
T.~B. Brown, B.~Mann, N.~Ryder, M.~Subbiah, J.~Kaplan, P.~Dhariwal, A.~Neelakantan, P.~Shyam, G.~Sastry, A.~Askell, S.~Agarwal, A.~Herbert{-}Voss, G.~Krueger, T.~Henighan, R.~Child, A.~Ramesh, D.~M. Ziegler, J.~Wu, C.~Winter, C.~Hesse, M.~Chen, E.~Sigler, M.~Litwin, S.~Gray, B.~Chess, J.~Clark, C.~Berner, S.~McCandlish, A.~Radford, I.~Sutskever, and D.~Amodei, ``Language models are few-shot learners,'' in \emph{Proc. Conf. Adv. Neural Inf. Process. Syst. (NeurIPS)}, virtual, Dec. 2020.

\bibitem{Chen2021decision}
L.~Chen, K.~Lu, A.~Rajeswaran, K.~Lee, A.~Grover, M.~Laskin, P.~Abbeel, A.~Srinivas, and I.~Mordatch, ``Decision transformer: Reinforcement learning via sequence modeling,'' in \emph{Proc. Conf. Adv. Neural Inf. Process. Syst. (NeurIPS) 6-14, 2021, virtual}, Dec. 2021, pp. 15\,084--15\,097.

\bibitem{mcmahan2017communication}
B.~McMahan, E.~Moore, D.~Ramage, S.~Hampson, and B.~A. y~Arcas, ``Communication-efficient learning of deep networks from decentralized data,'' in \emph{Proc. Int. Conf. Artif. Intell. Statist. (AISTATS)}, Ft. Lauderdale, FL, USA, Apr. 2017, pp. 1273--1282.

\bibitem{8824804}
S.~Yang, Y.~He, and X.~Zheng, ``{FoVR}: Attention-based {VR} streaming through bandwidth-limited wireless networks,'' in \emph{IEEE Int. Conf. Sens. Commun. Netw. (SECON)}, Boston, MA, USA, Jun. 2019, pp. 1--9.

\bibitem{9018228}
X.~Chen, A.~T.~Z. Kasgari, and W.~Saad, ``Deep learning for content-based personalized viewport prediction of 360-degree {VR} videos,'' \emph{IEEE Netw. Lett.}, vol.~2, no.~2, pp. 81--84, Feb. 2020.

\bibitem{10144339}
H.~Du, J.~Liu, D.~Niyato, J.~Kang, Z.~Xiong, J.~Zhang, and D.~I. Kim, ``Attention-aware resource allocation and {QoE} analysis for metaverse {xURLLC} services,'' \emph{IEEE J. Sel. Areas Commun.}, vol.~41, no.~7, pp. 2158--2175, Jun. 2023.

\bibitem{Saurabh2017FMADRL}
S.~Kumar, P.~Shah, D.~Hakkani{-}T{\"{u}}r, and L.~P. Heck, ``Federated control with hierarchical multi-agent deep reinforcement learning,'' [Online]. Available: https://arxiv.org/pdf/1712.08266.pdf.

\bibitem{JinPYWZ22FRL}
H.~Jin, Y.~Peng, W.~Yang, S.~Wang, and Z.~Zhang, ``Federated reinforcement learning with environment heterogeneity,'' in \emph{Proc. Int. Conf. Artif. Intell. Statist. (AISTATS)}, G.~Camps{-}Valls, F.~J.~R. Ruiz, and I.~Valera, Eds., vol. 151, Virtual Event, Mar. 2022, pp. 18--37.

\bibitem{KhodadadianSJM22}
S.~Khodadadian, P.~Sharma, G.~Joshi, and S.~T. Maguluri, ``Federated reinforcement learning: Linear speedup under markovian sampling,'' in \emph{Proc. Int. Conf. Mach. Learn. (ICML)}, ser. Proceedings of Machine Learning Research, K.~Chaudhuri, S.~Jegelka, L.~Song, C.~Szepesv{\'{a}}ri, G.~Niu, and S.~Sabato, Eds., vol. 162, Baltimore, Maryland, {USA}, Jul. 2022, pp. 10\,997--11\,057.

\bibitem{FanMDJTL21}
F.~X. Fan, Y.~Ma, Z.~Dai, W.~Jing, C.~Tan, and B.~K.~H. Low, ``Fault-tolerant federated reinforcement learning with theoretical guarantee,'' in \emph{Proc. Conf. Adv. Neural Inf. Process. Syst. (NeurIPS)}, M.~Ranzato, A.~Beygelzimer, Y.~N. Dauphin, P.~Liang, and J.~W. Vaughan, Eds., Virtual Event, Dec. 2021, pp. 1007--1021.

\bibitem{LiuWL19FRL}
B.~Liu, L.~Wang, and M.~Liu, ``Lifelong federated reinforcement learning: {A} learning architecture for navigation in cloud robotic systems,'' in \emph{IEEE/RSJ Int. Conf. Intell. Robot. Syst. (IROS)}, Macau, SAR, China, Nov. 2019, pp. 1688--1695.

\bibitem{WangLWLTL21}
X.~Wang, R.~Li, C.~Wang, X.~Li, T.~Taleb, and V.~C.~M. Leung, ``Attention-weighted federated deep reinforcement learning for device-to-device assisted heterogeneous collaborative edge caching,'' \emph{{IEEE} J. Sel. Areas Commun.}, vol.~39, no.~1, pp. 154--169, Dec. 2021.

\bibitem{YuCZGW21}
S.~Yu, X.~Chen, Z.~Zhou, X.~Gong, and D.~Wu, ``When deep reinforcement learning meets federated learning: Intelligent multitimescale resource management for multiaccess edge computing in 5g ultradense network,'' \emph{{IEEE} Internet Things J.}, vol.~8, no.~4, pp. 2238--2251, Mar. 2021.

\bibitem{xu2022PromptDT}
M.~Xu, Y.~Shen, S.~Zhang, Y.~Lu, D.~Zhao, J.~B. Tenenbaum, and C.~Gan, ``Prompting decision transformer for few-shot policy generalization,'' in \emph{Proc. Int. Conf. Mach. Learn. (ICML)}, ser. Proceedings of Machine Learning Research, vol. 162.\hskip 1em plus 0.5em minus 0.4em\relax Baltimore, Maryland, {USA}: {PMLR}, Jul. 2022, pp. 24\,631--24\,645.

\bibitem{ShridharTGBHMZF20}
M.~Shridhar, J.~Thomason, D.~Gordon, Y.~Bisk, W.~Han, R.~Mottaghi, L.~Zettlemoyer, and D.~Fox, ``{ALFRED:} {A} benchmark for interpreting grounded instructions for everyday tasks,'' in \emph{Proc. IEEE/CVF Conf. Comput. Vision Pattern Recognit. (CVPR)}.\hskip 1em plus 0.5em minus 0.4em\relax Seattle, WA, USA: Computer Vision Foundation / {IEEE}, Jun. 2020, pp. 10\,737--10\,746.

\bibitem{Seunghyun2023ClinicalDT}
S.~Lee, D.~Y. Lee, S.~Im, N.~H. Kim, and S.~Park, ``Clinical decision transformer: Intended treatment recommendation through goal prompting,'' [Online]. Available: \url{https://arxiv.org/pdf/2302.00612.pdf}.

\bibitem{shao2023survey}
J.~Shao, Z.~Li, W.~Sun, T.~Zhou, Y.~Sun, L.~Liu, Z.~Lin, and J.~Zhang, ``A survey of what to share in federated learning: Perspectives on model utility, privacy leakage, and communication efficiency,'' [Online]. Available: \url{https://arxiv.org/pdf/2307.10655.pdf}.

\bibitem{li2020federated}
T.~Li, A.~K. Sahu, M.~Zaheer, M.~Sanjabi, A.~Talwalkar, and V.~Smith, ``Federated optimization in heterogeneous networks,'' in \emph{Proc. Mach. Learn. Syst. ({MLSys})}, Austin, TX, USA, Mar. 2020.

\bibitem{zhou2022fedfa}
T.~Zhou, J.~Zhang, and D.~H.~K. Tsang, ``{FedFA}: Federated learning with feature anchors to align feature and classifier for heterogeneous data,'' \emph{IEEE Trans. Mobile Comput.}, pp. 1--17, Oct. 2023.

\bibitem{li2023fedcir}
Z.~Li, Z.~Lin, J.~Shao, Y.~Mao, and J.~Zhang, ``{FedCiR}: Client-invariant representation learning for federated non-iid features,'' [Online]. Available \url{https://arxiv.org/pdf/2308.15786.pdf}.

\bibitem{zhou2023understanding}
T.~Zhou, Z.~Lin, J.~Zhang, and D.~H.~K. Tsang, ``Understanding and improving model averaging in federated learning on heterogeneous data,'' [Online]. Available: \url{https://arxiv.org/pdf/2305.07845.pdf}.

\bibitem{Vaswani2017transformer}
A.~Vaswani, N.~Shazeer, N.~Parmar, J.~Uszkoreit, L.~Jones, A.~N. Gomez, L.~Kaiser, and I.~Polosukhin, ``Attention is all you need,'' in \emph{Proc. Conf. Adv. Neural Inf. Process. Syst. (NeurIPS)}, I.~Guyon, U.~von Luxburg, S.~Bengio, H.~M. Wallach, R.~Fergus, S.~V.~N. Vishwanathan, and R.~Garnett, Eds., Long Beach, CA, {USA}, Dec. 2017, pp. 5998--6008.

\bibitem{Xu2018video360}
Y.~Xu, Y.~Dong, J.~Wu, Z.~Sun, Z.~Shi, J.~Yu, and S.~Gao, ``Gaze prediction in dynamic 360{\textdegree} immersive videos,'' in \emph{Proc. IEEE/CVF Conf. Comput. Vision Pattern Recognit. (CVPR)}, Salt Lake City, UT, USA, Jun. 2018, pp. 5333--5342.

\bibitem{Reichl2010}
P.~Reichl, S.~Egger, R.~Schatz, and A.~D'Alconzo, ``The logarithmic nature of qoe and the role of the weber-fechner law in qoe assessment,'' in \emph{Proc. {IEEE} Int. Conf. Commun. (ICC)}, Cape Town, South Africa, May 2010, pp. 1--5.

\bibitem{9536410}
Z.~Gu, H.~Lu, P.~Hong, and Y.~Zhang, ``Reliability enhancement for {VR} delivery in mobile-edge empowered dual-connectivity sub-6 {GHz} and mmwave {HetNets},'' \emph{IEEE Trans. Wireless Commun.}, vol.~21, no.~4, pp. 2210--2226, Apr. 2022.

\bibitem{9174795}
W.~Sun, H.~Zhang, R.~Wang, and Y.~Zhang, ``Reducing offloading latency for digital twin edge networks in {6G},'' \emph{IEEE Trans. Veh. Technol.}, vol.~69, no.~10, pp. 12\,240--12\,251, Aug. 2020.

\bibitem{si2022enabling}
T.~S. Salem, G.~Iosifidis, and G.~Neglia, ``Enabling long-term fairness in dynamic resource allocation,'' \emph{Proc. {ACM} Meas. Anal. Comput. Syst.}, vol.~6, no.~3, pp. 46:1--46:36, Jan. 2021.

\bibitem{7588099}
T.~Hoßfeld, L.~Skorin-Kapov, P.~E. Heegaard, and M.~Varela, ``Definition of {QoE} fairness in shared systems,'' \emph{IEEE Commun. Lett.}, vol.~21, no.~1, pp. 184--187, Oct. 2017.

\bibitem{Matheron0S20Understanding}
G.~Matheron, N.~Perrin, and O.~Sigaud, ``Understanding failures of deterministic actor-critic with continuous action spaces and sparse rewards,'' in \emph{Artif. Neural Netw. and Mach. Learn. (ICANN)}, ser. Lecture Notes in Computer Science, I.~Farkas, P.~Masulli, and S.~Wermter, Eds., vol. 12397, Sep. 2020, pp. 308--320.

\bibitem{SchickS21SmallLanguage}
T.~Schick and H.~Sch{\"{u}}tze, ``It's not just size that matters: Small language models are also few-shot learners,'' in \emph{Proc. North Am. Rev. of Assoc. Comput. Linguist. (NAACL)}, Virtual Event, Jun. 2021, pp. 2339--2352.

\end{thebibliography}

\end{document}